\documentclass[10pt,twocolumn,letterpaper]{article}

\usepackage[pagenumbers]{iccv} 

\usepackage{graphicx}
\usepackage{makecell}
\usepackage{multicol}
\usepackage{colortbl}
\usepackage{multirow}

\definecolor{iccvblue}{rgb}{0.21,0.49,0.74}
\usepackage[pagebackref,breaklinks,colorlinks,allcolors=iccvblue]{hyperref}

\def\paperID{8320} 
\def\confName{ICCV}
\def\confYear{2025}

\title{FALCON: Resolving Visual Redundancy and Fragmentation in High-resolution Multimodal Large Language Models via Visual Registers}

\author{
    \textbf{\makecell{Renshan Zhang$^{1}$, Rui Shao$^{1}$\footnotemark[2], Gongwei Chen$^{1}$, Miao Zhang$^{1}$, \\ Kaiwen Zhou$^{2}$,  Weili Guan$^{1}$, Liqiang Nie$^{1}$\footnotemark[2]}}\\
    $^{1}$Harbin Institute of Technology, Shenzhen\quad
    $^{2}$Huawei Noah's Ark Lab\\
    \texttt{\normalsize{zhangrenshan@stu.hit.edu.cn}}\hspace{0.5cm}\texttt{\normalsize{shaorui@hit.edu.cn}}\\\texttt{\normalsize{ chengongwei@hit.edu.cn}}\hspace{0.5cm}\texttt{\normalsize{nieliqiang@gmail.com}}\\
    \texttt{\normalsize{\url{https://github.com/JiuTian-VL/JiuTian-FALCON}}}
}

\begin{document}
\maketitle

\renewcommand{\thefootnote}{\fnsymbol{footnote}} 
\footnotetext[2]{Corresponding authors}

\begin{abstract}

The incorporation of high-resolution visual input equips multimodal large language models (MLLMs) with enhanced visual perception capabilities for real-world tasks. However, most existing high-resolution MLLMs rely on a cropping-based approach to process images, which leads to fragmented visual encoding and a sharp increase in redundant tokens. To tackle these issues, we propose the \textbf{FALCON} model. FALCON introduces a novel \textbf{visual register} technique to simultaneously: \textbf{1)} \textbf{Eliminate redundant tokens at the stage of visual encoding.} To directly address the visual redundancy present in the output of vision encoder, we propose a Register-based Representation Compacting (\textbf{ReCompact}) mechanism. This mechanism introduces a set of learnable visual registers designed to adaptively aggregate essential information while discarding redundancy. It enables the encoder to produce a more compact visual representation with a minimal number of output tokens, thus eliminating the need for an additional compression module.
\textbf{2)} \textbf{Ensure continuity in visual encoding.} To address the potential encoding errors caused by fragmented visual inputs, we develop a Register Interactive Attention (\textbf{ReAtten}) module. This module facilitates effective and efficient information exchange across sub-images by enabling interactions between visual registers. It ensures the continuity of visual semantics throughout the encoding.
We conduct comprehensive experiments with FALCON on high-resolution benchmarks across a wide range of scenarios. FALCON demonstrates superior performance with a remarkable \textbf{9-fold} reduction in visual tokens.

\end{abstract}    
\vspace*{-7mm}
\section{Introduction}
\label{sec:intro}

\begin{figure}[t] 
    \centering
    \includegraphics[width=1\linewidth]{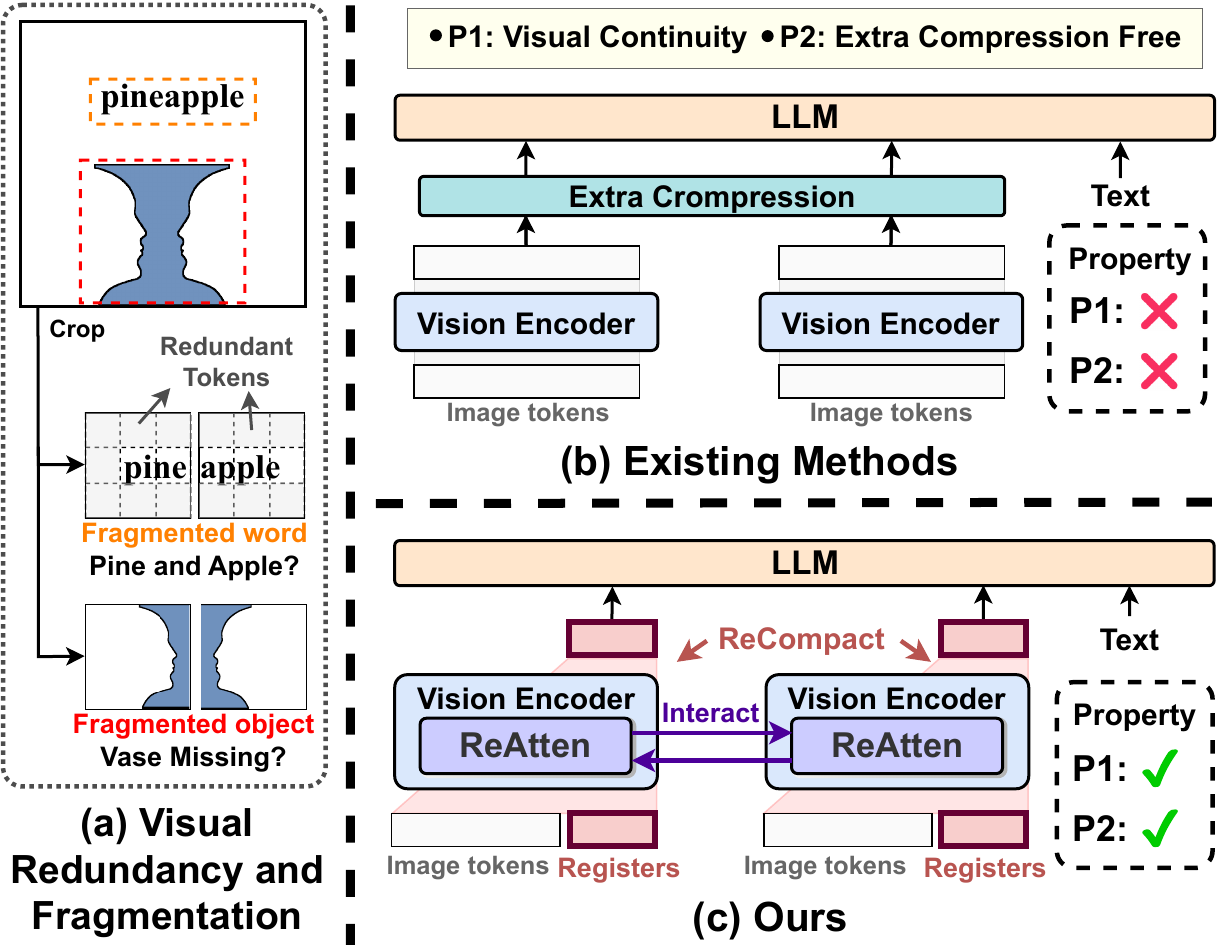}
    \caption{Comparison between existing high-resolution MLLMs and FALCON. FALCON eliminates visual redundancy and ensures the continuity of high-resolution encoding through the use of ReCompact and ReAtten mechanisms.}
    \label{fig:intro}
\vspace*{-7mm}
\end{figure}

Recently, multimodal large language models (MLLMs)~\cite{liu2024visual,liu2024improved,chen2024lion,ye2023mplug,ye2024mplug-owl2,dai2023instructblip} have made significant advancements in performing various vision-language tasks~\cite{shao2023detecting,shao2024detecting,yu2016modeling}, facilitating more natural human-machine interactions. Early MLLMs~\cite{liu2024visual,ye2023mplug,dai2023instructblip} typically processed images at a fixed low resolution, such as 224×224. This limited their capabilities in many real-world applications that require high-resolution input. To equip MLLMs with high-resolution visual perception, recent efforts~\cite{liu2024textmonkey,ye2023ureader, hu2024mplug_docowl_1_5,zhang2024beyond,huang2024hires} have introduced a cropping-based high-resolution processing method. Specifically, As shown in~\cref{fig:intro}(b), this approach crops high-resolution images into multiple non-overlapping sub-images that match the pre-trained resolution of the encoder. Each sub-image is processed independently by the vision encoder. Then the resulting visual features are concatenated and fed into the large language model (LLM).

Although effective, cropping-based methods still face issues of: \textbf{1)} \textbf{Visual Redundancy}.  The number of visual tokens grows dramatically with increasing resolution. While this enriches the visual details, it also leads to an increase in redundant tokens. These redundant tokens significantly increase the computational burden on LLM and interfere with the processing of visual information~\cite{zhang2024redundancy}.
To address these issues, as illustrated in~\cref{fig:intro}(b), existing methods typically rely on additional compression techniques. Representative approaches include combining learnable queries with cross-attention~\cite{bai2023qwen,ye2024mplug-owl2,li2024monkey} or simply employing pooling layers~\cite{yao2024deco,shi2024when_do}. However, these methods either demand substantial training resources or fail to fully address redundancy.
\textbf{2)} \textbf{Visual Fragmentation}. Cropping an image into multiple non-overlapping sub-images may disrupt visual semantics and potentially lead to encoding errors. As shown in~\cref{fig:intro}(a), in an OCR task, the word ``pineapple" may be incorrectly segmented into ``pine" and ``apple", causing the LLM to misinterpret the word. Similarly, fragmentation can also hinder the identification of objects. For example, the ``rubin vase" pattern illustrated in~\cref{fig:intro}(a) depicts both a vase and two faces simultaneously. However, the fragmentation could lead to the vase being unrecognizable.

To address these issues, as shown in~\cref{fig:intro}(c), we propose the \textbf{FALCON} model. FALCON introduces a novel \textbf{visual register} technique to simultaneously:
\textbf{1)} \textbf{Eliminate redundant tokens during visual encoding}. To eliminate redundancy directly at the stage of visual encoding, we propose a novel Register-based Representation Compacting (\textbf{ReCompact}) mechanism. This mechanism introduces a set of learnable visual registers as additional inputs to the vision encoder. The registers are designed to adaptively aggregate essential information from image tokens and discard redundant content. Then the outputs corresponding to the visual registers are considered as a non-redundant visual representation, serving as the compact input for the LLM instead of a large number of image tokens. This mechanism enables the vision encoder to convey comprehensive visual information in a compact representation using minimal tokens, thus removing the need for an additional compression module.
\textbf{2)} \textbf{Ensure continuity in visual encoding}. To avoid the potential encoding errors caused by fragmented visual inputs, we develop a novel Register Interactive Attention (\textbf{ReAtten}) module.
This module leverages the compact and comprehensive representation produced by visual registers to facilitate information exchange across sub-images. It gathers visual registers from different sub-images and employs a specialized cross-ViT attention module to enable interactions among them. Such interaction enables thorough information exchange across the entire image, thus ensuring visual continuity throughout the encoding process. Additionally, the compactness of the visual registers further enhances the efficiency of these interactions. Consequently, this module provides LLMs with a reliable foundation to accurately interpret visual content and reduce hallucinations.

Note that the widely used Abstractor-like modules~\cite{ye2024mplug-owl2, ye2023ureader, liu2024textmonkey, li2024monkey}, including QFormer~\cite{dai2023instructblip}, ReSampler~\cite{bai2023qwen}, follow a similar approach by using query tokens to extract information from image tokens after visual encoding. The key difference is that the visual register mechanism directly adopts the pretrained ViT to aggregate information into a set of specified register tokens. This design is based on the observation that a well-pretrained ViT can naturally gather global information into some specific tokens through self-attention~\cite{darcet2024vision}. It offers the visual register mechanism several strengths in terms of:
\textbf{1) Training Efficiency}. The substantial number of randomly initialized parameters in Abstractor-like modules requires extensive data for pretraining. For instance, QFormer~\cite{li2023blip2}, Abstractor~\cite{ye2024mplug-owl2}, and Resampler~\cite{bai2023qwen} use large corpora of 129M, 400M, and 1.4B samples, respectively. In contrast, by leveraging well-pretrained ViT parameters, the visual register mechanism requires fewer than 3M samples for adaptation. The ablation results in~\cref{fig:ablation_recompact} further demonstrate this advantages.
\textbf{2) Addressing Visual Fragmentation}. By leveraging the compact and comprehensive representation produced by visual registers, the proposed ReAtten module can effectively facilitate information exchange across sub-images. This is an important function that is notably lacking in Abstractor-like modules. \textbf{Combined with ReCompact, it enables the visual register mechanism to simultaneously address the issues of visual redundancy and fragmentation.} The ablation results in~\cref{tab:ablation_fragmentation} validate the effectiveness of ReAtten.

We summarize our main contributions as follows:
\begin{itemize}[leftmargin=*]
    \item
    To address the issue of visual redundancy directly at the encoding phase, we introduce the Register-based Representation Compacting (ReCompact) mechanism. This mechanism enables vision encoder to capture rich visual information using only a minimal number of tokens, effectively eliminating redundancy and obviating the need for additional compression modules.
    \item 
    To address the issue of visual fragmentation caused by image cropping, we propose the register interactive attention (ReAtten) module. This module ensures the semantic continuity of high-resolution visual encoding in an efficient and effective manner.
    \item
    We introduce FALCON, a model that leverages the proposed visual register technique. Comprehensive evaluations on high-resolution visual understanding benchmarks highlight the superiority of FALCON. Notably, FALCON achieves exceptional results with a remarkable reduction in visual tokens by 9×.
\end{itemize}

\begin{figure*}[t] 
    \centering
    \includegraphics[width=1\linewidth]{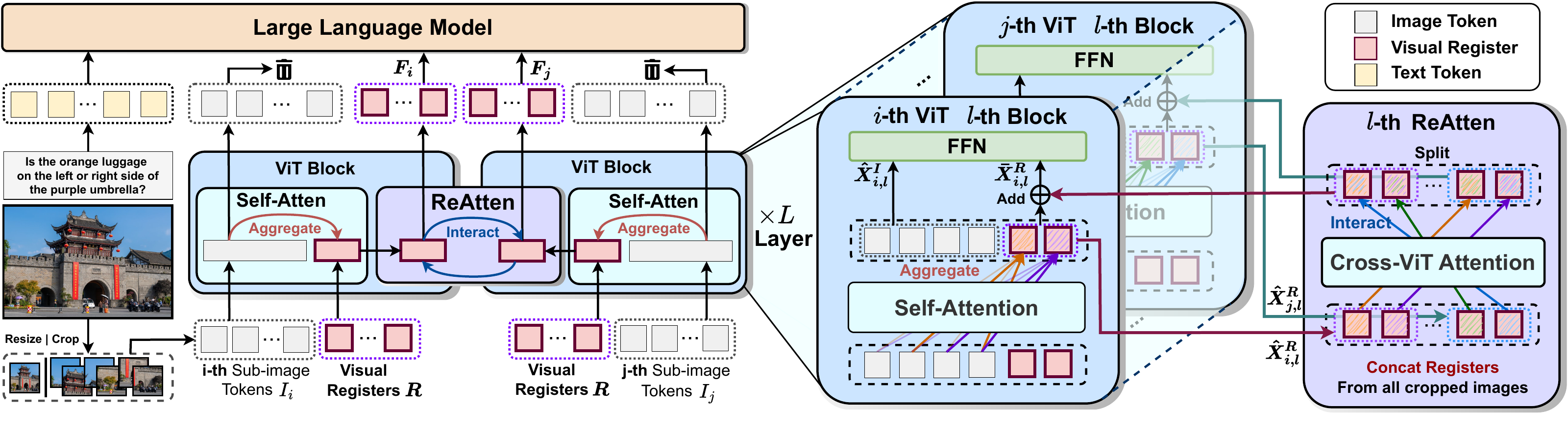}
    \caption{Overall framework of the proposed FALCON. The model introduces a set of learnable visual registers, which are fed into the ViT along with the image tokens from each sub-image. During the visual encoding, the ReCompact mechanism enables information from the image tokens to be aggregated into the visual registers, while the proposed ReAtten facilitates information exchange among the visual registers across different sub-images. Ultimately, these visual registers provide a compact visual representation as input to the LLM.}
    \label{fig:arch}
    \vspace*{-5mm}
\end{figure*}

\section{Related Work}
\label{sec:related_work}

\subsection{Multimodal Large Language Models}
Research on MLLMs~\cite{liu2024improved,chen2024lion,bai2023qwen,ye2023mplug,ye2024mplug-owl2,shen2024mome} has made significant strides in recent years. By incorporating image inputs into pretrained LLMs~\cite{touvron2023llama,touvron2023llama2,chiang2023vicuna}, MLLMs demonstrate strong capabilities across various vision-language tasks. Typically, these models use a pretrained vision encoder~\cite{radford2021learning,zhai2023sigmoid} to transform input images into a set of visual tokens. These tokens are then projected into the input space of the LLM through a projection module (e.g., MLP~\cite{liu2024visual,liu2024improved}, QFormer~\cite{dai2023instructblip} and Abstractor~\cite{ye2024mplug-owl2}) and processed alongside text inputs. 
Building on the progress of MLLMs, recent studies have shifted from designing specialized methods for certain tasks~\cite{shao2023detecting,shao2024detecting,shao2019multi} to employing MLLMs as generalist agents, capable of handling diverse practical tasks such as document analysis~\cite{ye2023mplug_docowl,hu2024mplug_docowl_1_5,zhang2024token}, online video assistant~\cite{li2025lion}, game agents~\cite{li2024optimus,li2025optimus2}, robot manipulation~\cite{li2025star,kimopenvla,lv2025spatial}, and GUI agent~\cite{chen2025spabench,chen2025SimpAgent,xie2025gui}.
Despite these advances, MLLMs still struggle with the fixed low resolution of pretrained image encoders, which can limit their effectiveness in handling high-resolution images.

\subsection{High-resolution MLLMs}

To address the limitation of MLLMs in handling only low-resolution visual inputs, early studies attempted to either directly incorporate vision encoders that were pre-trained on higher-resolution images~\cite{lu2024deepseek,luo2024feast}, or to improve the resolution of pre-trained encoders through additional training~\cite{bai2023qwen,ye2024mplug-owl2}. While achieving some progress, they still remain restricted by the fixed resolutions. To accommodate the varying resolutions and aspect ratios, cropping-based techniques have been widely adopted in recent works~\cite{ye2023ureader,hu2024mplug_docowl_1_5,huang2024hires,zhang2024beyond,liu2024textmonkey}. However, these approaches still face challenges related to visual redundancy and fragmentation.

To mitigate the issue of redundancy, existing models typically require an additional compression module. For example, Abstractor-like modules utilizing learnable queries combined with cross-attention are widely adopted~\cite{bai2023qwen,ye2024mplug-owl2,ye2023ureader,li2024monkey,liu2024textmonkey}. These modules typically require pretraining on massive amounts of data, which diminishes their practical efficacy. In contrast, some approaches use pooling layers~\cite{yao2024deco, shi2024when_do} or convolutional layers~\cite{lu2024deepseek,hu2024mplug_docowl_1_5}.
Though resource-efficient, they often fail to address the issue of redundancy.
In this work, we present an initial exploration to directly eliminate redundancy during the visual encoding phase, thereby removing the need for additional compression. To address the issue of fragmentation, several concurrent studies also make preliminary investigation. TextMonkey~\cite{liu2024textmonkey} uses Shifted Window Attention~\cite{liu2021swin} to exchange information between sub-images. But the exchange is still limited to local regions. MiniMonkey~\cite{huang2024minimonkey} proposed a Complementary Image Pyramid strategy to generate complementary cropping schemes. However, this strategy inevitably encodes the same visual content multiple times, further exacerbating visual redundancy.
In this work, we propose the ReAtten module to facilitate continuous visual encoding through the interaction among visual registers.

\section{FALCON}
\label{sec:method}

The overall framework of FALCON is illustrated in~\cref{fig:arch}. The method begins with shape-adaptive cropping~\cite{ye2023ureader}, where the image is cropped into multiple non-overlapping sub-images, each matching the resolution of the pre-trained vision encoder. Additionally, a resized version of the full image is used to supplement global layout information.
To directly eliminate redundancy during visual encoding, we propose Register-based Representation Compacting (ReCompact) mechanism. This mechanism introduces a set of learnable visual registers, which are paired with image tokens from each sub-image and fed into the vision encoder to capture rich visual information. To ensure the continuity of visual semantics throughout the encoding, a novel Register Interactive Attention (ReAtten) module is integrated into the Vision Transformer (ViT)~\cite{dosovitskiy2021vit} to facilitate information exchange between sub-images via the visual registers. Finally, the compact visual representations produced by the visual registers are processed through a simple MLP module before being fed into the LLM for further analysis.

\subsection{Mitigating Visual Redundancy via ReCompact}
\label{sec:method_reg}

Existing MLLMs typically use a ViT-based vision encoder, such as CLIP~\cite{radford2021learning}. For an input image of size $H \times W$, it is patchified into multiple $P \times P$ patches. Each patch is treated as an individual image token, resulting in $N=\frac{H}{P} \times \frac{W}{P}$ tokens, denoted as $I=\{i_1,\cdots, i_N\}$. In most current MLLMs, output features corresponding to all image tokens are usually preserved in their entirety.
However, when high-resolution images are included, the number of image tokens increases dramatically, placing a substantial burden on the computational resources of LLMs. Consequently, most high-resolution MLLMs necessitate an additional compression module, which can be resource-inefficient. Previous research has shown that most image tokens contain only localized information from their surrounding areas~\cite{darcet2024vision}. It suggests that a significant portion of these tokens may be redundant, particularly those originating from background regions. This observation motivates us to directly eliminate redundancy in the output of the vision encoder, thereby reducing the number of visual tokens and avoiding the need for an additional compression module. 

To achieve this, we propose a novel Register-based Representation Compacting (\textbf{ReCompact}) mechanism. This mechanism employs a set of visual registers to aggregate essential information from image tokens, resulting in a more compact visual representation. Specifically, as shown in~\cref{fig:arch}, we introduce a set of learnable visual registers, denoted by $R = \{r_1, \cdots, r_M\}$, where $M$ represents the number of visual registers, and $M\ll N$. Given a single input image, the image tokens are concatenated with visual registers to create a new input for ViT, represented as $V = \text{Concat}(I, R) = \{i_1, \cdots, i_N, r_1, \cdots, r_M\}$. During visual encoding, the self-attention mechanism in ViT enables the visual register to adaptively aggregate important information from image tokens while discarding redundant content. Denote $X_{k,l} \in \mathcal{R}^{(N+M) \times D}$ as the input hidden states of the $k$-th sub-image at the $l$-th layer, where $D$ is the dimensionality. The self-attention is formulated as:
\begin{equation}
\label{eq:self_atten}
\hat{X}_{k,l} = \text{Self-Atten}(X_{k,l}) = \text{Softmax}(\frac{X_{k,l}X_{k,l}^T}{\sqrt{D_{key}}})X_{k,l}
\end{equation}
Notably, we do not apply any attention masks for specific constraints. Our experiments show that pretraining on a modest amount of data is sufficient to effectively adapt the pretrained ViT to the proposed ReCompact, enabling it to aggregate essential information within the visual registers.
After visual encoding, The output features corresponding to the visual registers, denoted as $F=\{f_1,\cdots,f_M\}$, are then used as the visual representation of the image and fed into the LLM for various vision-language tasks.

To further adapt the ReCompact mechanism for cropping-based high-resolution image perception, we share the same set of visual registers as input for different sub-images. The input for the $k$-th sub-image is formulated as $V_k = \text{Concat}(I_k, R)$, where $k = 1, 2, \cdots, N_c$ and $N_c$ denotes the total number of sub-images. After encoding, the outputs of the visual registers for all sub-images are concatenated, yielding $F_{hr}=\text{Concat}(F_1, F_2, \cdots, F_{N_c})$, which serves as the input to the LLM. The ReCompact mechanism enables the vision encoder to represent rich visual information using only a minimal number of $M$ tokens per sub-image, which significantly reduces the visual redundancy and eliminates the need for additional compression.

\subsection{Ensuring Visual Continuity via ReAtten}
\label{sec:method_reatten}

When using the cropping-based high-resolution image encoding method, each sub-image is independently encoded by the vision encoder. This process can disrupt semantic coherence and result in the misrepresentation of visual concepts. To preserve the continuity of visual semantics, it is crucial to enable information exchange between sub-images during encoding. A naive approach would be to directly concatenate the image features from all sub-images and apply self-attention. However, the quadratic computational complexity makes this approach impractical. In~\cref{sec:method_reg}, we introduced the ReCompact mechanism, where the visual registers effectivly aggregate rich visual information while discarding redundancy during visual encoding, allowing them to serve as compact yet comprehensive image representations. Consequently, interactions between visual registers can naturally facilitate sufficient information exchange across different sub-images in an efficient manner. 

To enable information exchange between visual registers across different sub-images, we introduce a novel Register Interactive Attention (\textbf{ReAtten}) module. As shown in~\cref{fig:arch}, the ReAtten module is integrated into each of the $L$ layers of the ViT. After the self-attention process in Eq. (1), we acquire $\hat{X}_{k,l} = \{\hat{X}^I_{k,l}, \hat{X}^R_{k,l}\}$, where $\hat{X}^I\in \mathcal{R}^{N\times D}$ and $\hat{X}^R\in \mathcal{R}^{M\times D}$ represent the hidden states corresponding to image tokens and visual registers, respectively. The states of the visual registers from different sub-images are concatenated to form $\hat{X}^R_{l} = \text{Concat}\left(\{\hat{X}^R_{k,l}\}_{k=1}^{N_c}\right) \in \mathcal{R}^{M \cdot N_c \times D}$. Then these concatenated registers are processed via ReAtten to enable information exchange across sub-images, which can be formulated as:
\begin{equation}
\overline{X}^R_{l} = \text{ReAtten}(\hat{X}^R_{l}) = \hat{X}^R_{l} + \text{Cross-ViT-Atten}(\hat{X}^R_{l})
\end{equation}
where Cross-ViT-Atten is implemented as a self-attention layer. Once the information exchange is completed, the registers and image tokens from a same sub-image are concatenated and then passed into the feedforward network (FFN) for transformation, yielding the output of the $l$-th layer:

\begin{equation}
    X_{k,l+1}=\text{FFN}\left(\text{Concat}(\hat{X}^I_{k,l},\overline{X}^R_{k,l})\right)
\end{equation}

By integrating ReAtten into each layer of the ViT model, FALCON enables effective information exchange between cropped sub-images through visual registers, thereby ensuring semantic continuity of the high-resolution image throughout the visual encoding process.

\begin{figure}[t] 
    \centering
    \includegraphics[width=0.9\linewidth]{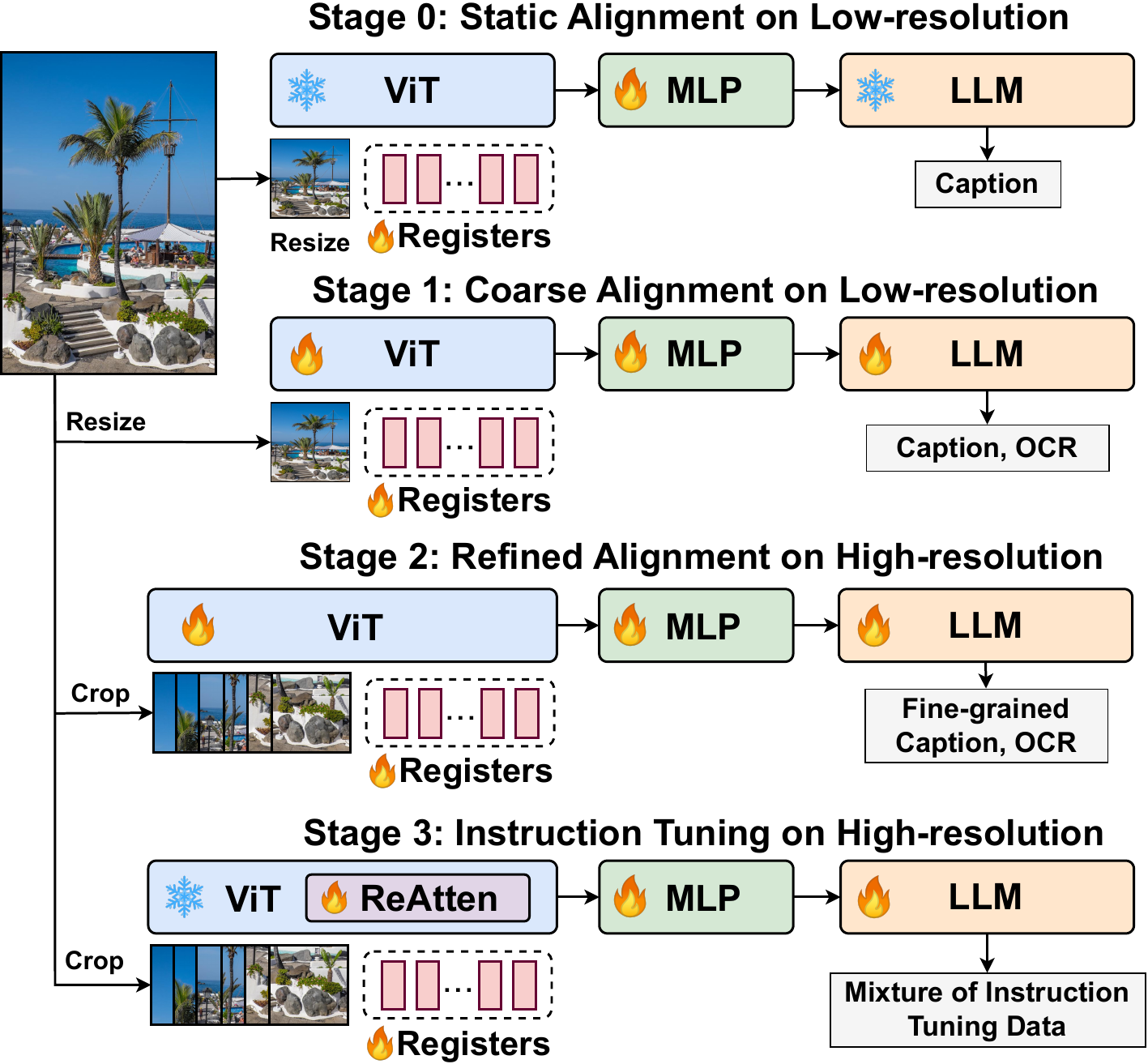}
    \caption{The progressive training pipeline. \textbf{Stage 0}: Align the frozen vision and language backbones by training only the MLP and registers. \textbf{Stage 1}: Tune all parameters with caption and OCR data on low-resolution for coarse alignment. \textbf{Stage 2}: Integrate high-resolution input along with fine-grained captions and OCR to achieve refined alignment. \textbf{Stage 3}: Froze ViT and introduce ReAtten module for high-resolution instruction tuning.}
    \label{fig:pipeline}
    \vspace*{-5mm}
\end{figure}

\subsection{Learning Expressive Visual Representation via Progressive Training Pipeline}
\label{sec:method_pipeline}

To empower the ReCompact mechanism with the capability to capture rich visual patterns and to enable smoother initialization of the ReAtten module under high-resolution conditions, we develop a specialized progressive training pipeline. As illustrated in~\cref{fig:pipeline}, this pipeline progressively transitions from low to high resolution across four stages.

\textbf{Stage 0: Static Alignment on Low-resolution}. Since both the visual registers and MLP projector are randomly initialized, in stage 0, we aim to achieve a initial alignment between the pretrained vision encoder and the LLM. To facilitate this, we begin by using low-resolution images as inputs, which are directly resized to match the resolution of the vision encoder. We then statically pretrain the model by keeping both the vision encoder and LLM frozen, tuning only the visual registers and MLP using image caption data.

\textbf{Stage 1: Coarse Alignment on Low-resolution}. To better adapt the pre-trained vision encoder to our newly introduced ReCompact mechanism, in stage 1, we unfreeze all model parameters to enable a more thorough tuning. To facilitate the visual registers to progressively learn capturing visual information from coarse to fine granularity, we continue to use low-resolution inputs and employ high-quality, extended captions rich in visual content, along with full-text OCR data for training. This design allows the registers to acquire holistic scene understanding ability.

\textbf{Stage 2: Refined Alignment on High-resolution}. 
To further enhance the capacity of the visual register to capture fine-grained visual details, in stage 2,  we transition the model to process high-resolution image for refined alignment.
Additionally, we employ datasets that includes a series of fine-grained captions and OCR tasks, such as regional captions and text location. This facilitates the visual registers to capture comprehensive fine-grained details.

\textbf{Stage 3: Instruction Tuning on High-resolution}. To adapt the well-aligned visual register for a wide range of real-world tasks, in stage 3, we perform instruction tuning with a mixture of datasets covering various tasks, such as VQA, document understanding, and science reasoning. Additionally, the proposed ReAtten module is integrated to address visual fragmentation caused by cropping-based high-resolution image encoding. Since the vision encoder with ReCompact is well-aligned, the vision encoder is kept frozen and other parameters including ReAtten are trainable during instruction tuning. Notably, to facilitate a smoother initialization for the newly introduced ReAtten modules, we initialize parameters of each ReAtten module with the ViT self-attention parameters at the corresponding depth. This design enables ReAtten to effectively leverage pre-trained parameters, avoiding the need for extensive fine-tuning.
\section{Experiments}
\label{sec:experiments}


We implement FALCON with SigLIP-L/16-384px~\cite{zhai2023sigmoid} as the vision encoder and Llama-3.1-8B-Instruct~\cite{dubey2024llama3} as the LLM backbone. A two-layer MLP module with GeLU~\cite{hendrycks2016bridging} activation is served as the vision-language projection layer. The number of visual registers is set to 64, achieving remarkable visual token compression rates of $9\times$. The maximum number of sub-images is set to 16. 

\subsection{Experimental Results}

\begin{table*}[t]
\small
\caption{Evaluation on MME-RealWorld Benchmark. ``OCR", ``RS", ``DT", ``MO", and ``AD" stand for OCR in the Wild, Remote Sensing, Diagram and Table, Monitoring, and Autonomous Driving, respectively.
``Avg" denotes the weighted average accuracy across all samples, while ``Avg-C" denotes the weighted average accuracy across different domains.
``$\text{Size}^v$" indicates the image resolution supported by the model, where $\times n$ signifies a cropping-based high-resolution processing approach with up to $n$ sub-images. ``$\text{Token}^c$" denotes the number of visual tokens after compression. \textdagger~For models utilizing hybrid vision encoders, ``$\text{Size}^v$" represents the setting of encoder with the highest resolution. We \textcolor{gray!90}{gray} out the models that are either closed-source or have at least \textbf{twice} the number of visual tokens compared to ours. Among these, the lighter \textcolor{gray!50}{gray} denotes scores below ours. The \underline{underline} indicates that the second performance is achieved. Notably, FALCON achieves state-of-the-art performance with just \textbf{64} tokens per sub-image, surpassing most models that have large amount of visual tokens and many advanced closed-source commercial models.}
\label{tab:mme_realword}
\centering
\setlength{\tabcolsep}{1.5pt}
\begin{tabular}{c | l cc| ccccc|cc| cccc|cc}
    \toprule
     Type & \multicolumn{1}{c}{\textbf{Method}} & \multicolumn{1}{c}{\textbf{$\text{Size}^v$}} & \multicolumn{1}{c}{\textbf{$\text{Token}^c$}} & \multicolumn{7}{|c|}{\textbf{Perception}}  & \multicolumn{6}{c}{\textbf{Reasoning}}\\
    
    \midrule 
     & \multicolumn{3}{c|}{\textbf{Task Split}} & \textbf{OCR} & \textbf{RS} & \textbf{DT} & \textbf{MO} & \textbf{AD} & \hphantom{-}\textbf{Avg}\hphantom{-}& \textbf{Avg-C} & \textbf{OCR} & \textbf{DT} & \textbf{MO} & \textbf{AD} & \hphantom{-}\textbf{Avg}\hphantom{-} & \textbf{Avg-C} \\ 

    \midrule

     \multirow{10}{*}{\rotatebox{90}{\makecell{Visual Tokens\\$\leq$ 1024}}} & MiniGPT-v2~\cite{chen2023minigpt} & $448^2$ &  $256$ & 39.0 & 23.3 & 20.4 & 19.3 & 26.0 & {26.9} & {25.6} & 30.0 & 20.4 & 16.9 & 23.7 & {23.0}& {22.7}\\

     & {Qwen-VL-Chat~\cite{bai2023qwen}} & {$448^2$} &  {$256$} & {32.4} & {15.1} & {15.6} & {22.1} & {15.1} & {20.8} & {20.1} & {28.6} & {13.6} & {16.5} & {24.6} & {22.0} & {20.8} \\

     & LLaVA1.5-13B~\cite{liu2024improved}   &$336^2$ &  $576$ & 44.1 & 23.3 & 20.2 & 20.5 & 26.1 & {28.4} & {26.8} & 30.2 & 20.8 & 27.5 & 24.8 & {25.5}& {25.8}\\

     & ShareGPT4V-13B~\cite{Lin2023ShareGPT4V}  & $336^2$&  $576$ & 44.6 & 23.1 & 20.2 & 19.3 & 26.1 & {28.4} & {26.6}  & 26.0 & 20.8 & 27.3 & 24.6 & {24.6}& {24.7}\\

     & TextMonkey~\cite{liu2024textmonkey} & $448^2\times4$ &  $256\times4$ & 37.3 & 11.7 & 5.9 & 16.1 & 14.3 & {18.2} & {17.1} & 30.4 & 2.2 & 4.4 & 20.0 & {16.0}& {14.3} \\ 
     
     & Mini-Gemini-7B-HD\textdagger~\cite{li2024mini_gemini} & $336^2\times4$ &  $576$ & 42.0 & 31.3 & 22.3 & 34.2 & 24.8 & {31.1} &{30.9}& 35.4 & 24.6 & 25.9 & 23.3 & {26.1}& {27.3} \\

     & DeepSeek-VL\textdagger~\cite{lu2024deepseek} & $1024^2$ &  $576$ & 49.6 & 25.5 & 23.4 & 27.0 & 33.4 & {33.1} & {31.8} & 45.2 & 23.8 & 16.7 & 27.3 & {28.0} & {28.3} \\

     & {MiniCPM-V 2.5~\cite{yao2024minicpm}}  & {$448^2\times 9$} & {$96\times 9$} & \textbf{66.8} & {27.7} & \textbf{52.8} & {38.7} & {34.2} & {47.4} & {44.0} & {44.0} & {31.8} & {37.0} & {31.0} & {34.5}& {36.0} \\

     & Monkey~\cite{li2024monkey} & $448^2\times 4$ &  $256\times4$ & 54.6 & 25.0 & 32.5 & 28.0 & 29.7 & {36.3}& {34.0} & 27.2 & 20.8 & 27.3 & 33.0 & {28.8} &  {27.1} \\

    \rowcolor[HTML]{faf0e6}
     & FALCON & $384^2\times16$ &  $\textbf{64}\times16$ & \underline{66.4} & \textbf{49.6} & \underline{47.2} & \textbf{40.4} & \textbf{36.5} & \textbf{50.3} & \textbf{48.0} & \textbf{55.4} & \textbf{39.2} & \textbf{44.4} & \textbf{34.5} & \textbf{40.7}& \textbf{43.4}\\
    

     \midrule
     
     \multirow{6}{*}{\rotatebox{90}{\makecell{Visual Tokens\\$\geq$2048}}} & \textcolor{gray!90}{LLaVA-Next-Qwen72B~\cite{liu2024llavanext}} & \textcolor{gray!90}{$336^2\times4$} &  \textcolor{gray!90}{$576\times4$} & \textcolor{gray!50}{37.1} & \textcolor{gray!50}{29.1} & \textcolor{gray!50}{27.7} & \textcolor{gray!50}{29.4} & \textcolor{gray!50}{18.0} & \textcolor{gray!50}{29.0}& \textcolor{gray!50}{28.3} & \textcolor{gray!50}{17.2} & \textcolor{gray!50}{34.2} & \textcolor{gray!50}{27.3} & \textcolor{gray!50}{29.7} & \textcolor{gray!50}{27.9}& \textcolor{gray!50}{27.1} \\
     
     & \textcolor{gray!90}{LLaVA-Next-llama3~\cite{liu2024llavanext}} & \textcolor{gray!90}{$336^2\times4$} &  \textcolor{gray!90}{$576\times4$} & \textcolor{gray!50}{47.9} & \textcolor{gray!50}{25.4} & \textcolor{gray!50}{26.6} & \textcolor{gray!50}{19.5} & \textcolor{gray!50}{18.7} & \textcolor{gray!50}{30.1} & \textcolor{gray!50}{27.6} &  \textcolor{gray!50}{55.2} & \textcolor{gray!50}{23.4} & \textcolor{gray!50}{21.1} & \textcolor{gray!50}{30.7} & \textcolor{gray!50}{32.1}& \textcolor{gray!50}{32.6} \\

     & \textcolor{gray!90}{mPLUG-DocOwl 1.5~\cite{hu2024mplug_docowl_1_5}} & \textcolor{gray!90}{$448^2\times9$} &  \textcolor{gray!90}{$256\times9$} & \textcolor{gray!50}{51.2} & \textcolor{gray!50}{23.7} & \textcolor{gray!50}{29.3} & \textcolor{gray!50}{25.0} & \textcolor{gray!50}{28.3} & \textcolor{gray!50}{33.7} & \textcolor{gray!50}{31.5} & \textcolor{gray!50}{42.6} & \textcolor{gray!50}{19.8} & \textcolor{gray!50}{20.5} & \textcolor{gray!50}{26.0} & \textcolor{gray!50}{26.9}& \textcolor{gray!50}{27.2} \\
     
     & \textcolor{gray!90}{CogVLM2-llama3-Chat~\cite{hong2024cogvlm2}}  & \textcolor{gray!90}{$1344^2$} & \textcolor{gray!90}{$2304$} & \textcolor{gray!90}{70.0} & \textcolor{gray!50}{28.8} & \textcolor{gray!90}{47.5} & \textcolor{gray!50}{33.7} & \textcolor{gray!50}{30.2} & \textcolor{gray!50}{45.8} &\textcolor{gray!50}{42.0} & \textcolor{gray!50}{54.0} & \textcolor{gray!50}{32.8} & \textcolor{gray!50}{41.2} & \textcolor{gray!50}{31.2} & \textcolor{gray!50}{37.3} & \textcolor{gray!50}{39.8} \\

     & \textcolor{gray!90}{IXC2.5~\cite{zhang2024ixc2_5}}  & \textcolor{gray!90}{$560^2\times 24$} & \textcolor{gray!90}{$400\times 24$} & \textcolor{gray!90}{69.3} & \textcolor{gray!50}{36.1} & \textcolor{gray!90}{{63.9}} & \textcolor{gray!50}{39.5} & \textcolor{gray!50}{33.6} & \textcolor{gray!90}{52.5} & \textcolor{gray!90}{48.5} & \textcolor{gray!50}{53.4} & \textcolor{gray!90}{{41.0}} & \textcolor{gray!50}{17.7} & \textcolor{gray!50}{30.0} & \textcolor{gray!50}{{33.9}}& \textcolor{gray!50}{35.5} \\

     & \textcolor{gray!90}{LLaVA-OneVision~\cite{li2024llava_onevision}}  & \textcolor{gray!90}{$384^2\times 9$} & \textcolor{gray!90}{$729\times 9$} & \textcolor{gray!90}{78.7} & \textcolor{gray!90}{53.5} & \textcolor{gray!90}{60.7} & \textcolor{gray!50}{40.3} & \textcolor{gray!90}{45.8} & \textcolor{gray!90}{59.6} & \textcolor{gray!90}{55.8} & \textcolor{gray!90}{61.8} & \textcolor{gray!90}{40.0} & \textcolor{gray!50}{40.8} & \textcolor{gray!50}{34.1} & \textcolor{gray!90}{41.2}& \textcolor{gray!90}{44.2} \\

    \midrule

    \multirow{4}{*}{\rotatebox{90}{\makecell{Closed-\\source}}} & \textcolor{gray!90}{GPT-4o-mini}  &- &  -& \textcolor{gray!50}{62.5} & \textcolor{gray!50}{6.7} & \textcolor{gray!50}{44.2} & \textcolor{gray!50}{26.5} & \textcolor{gray!50}{24.2} & \textcolor{gray!50}{{37.1}} & \textcolor{gray!50}{{32.8}} & \textcolor{gray!50}{47.0} & \textcolor{gray!90}{39.8} & \textcolor{gray!50}{25.8} & \textcolor{gray!50}{26.8} & \textcolor{gray!50}{{32.5}} & \textcolor{gray!50}{{34.9}}\\
     
     & \textcolor{gray!90}{Gemini-1.5-pro~\cite{team2023gemini}}   &- & - & \textcolor{gray!90}{67.6} & \textcolor{gray!50}{14.0} & \textcolor{gray!50}{39.9} & \textcolor{gray!50}{31.1} & \textcolor{gray!50}{26.6} & \textcolor{gray!50}{{39.6}} & \textcolor{gray!50}{{35.9}} & \textcolor{gray!50}{52.7} & \textcolor{gray!50}{33.2} & \textcolor{gray!50}{28.3} & \textcolor{gray!50}{19.2} & \textcolor{gray!50}{{29.2}}& \textcolor{gray!50}{{33.4}}\\
     
     & \textcolor{gray!90}{GPT-4o}   & -& - & \textcolor{gray!90}{{77.7}} & \textcolor{gray!50}{{28.9}} & \textcolor{gray!50}{46.7} & \textcolor{gray!50}{{33.9}} & \textcolor{gray!50}{22.4} & \textcolor{gray!50}{46.4} & \textcolor{gray!50}{{41.9}} & \textcolor{gray!90}{61.4} & \textcolor{gray!90}{44.8} & \textcolor{gray!50}{36.5} & \textcolor{gray!50}{26.4} & \textcolor{gray!50}{37.6} & \textcolor{gray!50}{42.3}\\
     
     & \textcolor{gray!90}{{Claude 3.5 Sonnet}}  &- & - & \textcolor{gray!90}{72.5} & \textcolor{gray!50}{25.7} & \textcolor{gray!90}{{67.4}} & \textcolor{gray!50}{32.2} & \textcolor{gray!90}{{40.8}} & \textcolor{gray!90}{{52.9}} & \textcolor{gray!50}{{47.7}}& \textcolor{gray!90}{61.9} & \textcolor{gray!90}{{61.2}} & \textcolor{gray!50}{{41.8}} & \textcolor{gray!50}{31.9} & \textcolor{gray!90}{44.1} & \textcolor{gray!90}{{49.2}}\\
    
    \bottomrule
    \end{tabular}%
\vspace*{-3mm}
\end{table*}


We evaluate FALCON using the recently proposed MME-RealWorld~\cite{zhang2024mme_realword} benchmark, a comprehensive and large-scale benchmark specifically designed to assess the high-resolution understanding capabilities of MLLMs. MME-RealWorld comprises 13,366 high-resolution images with an average resolution of 2,000×1,500 pixels, spanning five domains: OCR in the wild, Remote Sensing, Diagrams and Tables, Monitoring, and Autonomous Driving. This benchmark is well-suited for evaluating the practical high-resolution understanding capabilities of MLLMs.

As shown in~\cref{tab:mme_realword}, FALCON significantly outperform other advanced MLLMs in both perception and reasoning tasks. FALCON achieves state-of-the-art performance with only \textbf{64} tokens per sub-image, resulting in a remarkable visual compression rate of 9 times. In addition, FALCON also outperforms most models that have twice as many visual tokens. Specifically, FALCON outperforms CogVLM2-llama3-Chat~\cite{hong2024cogvlm2} in average scores on both perception and reasoning task, which uses 2304 tokens per-image.
It also surpasses IXC2.5~\cite{zhang2024ixc2_5} in average scores on the reasoning tasks, which has a maximun token counts of 9600.
Notably, FALCON achieves higher average scores in both perception and reasoning tasks compared to three advanced commercial models, GPT-4o, GPT-4o-mini and Gemini-1.5-pro~\cite{team2023gemini}, with it notably outperforming all commercial models in the sub-tasks of remote sensing and monitoring. These results demonstrate that the ReCompact and ReAtten technique proposed by FALCON effectively eliminate visual redundancy and ensure visual continuity, allowing comprehensive and detailed visual information to be conveyed with a highly compact visual representation.

To further assess FALCON's capabilities in high-resolution understanding, we conduct evaluations on more high-resolution benchmarks: V*~\cite{wu2024vstar}, DocVQA~\cite{mathew2021docvqa}, TextVQA~\cite{singh2019textvqa} and ChartQA~\cite{masry2022chartqa}. V* focuses on evaluating models' capabilities for fine-grained attribute recognition and spatial reasoning. DocVQA challenges models to comprehend text-dense document images. ChartQA evaluates models' proficiency in interpreting various types of chart, while TextVQA assesses the ability to recognize small text in real-world scenes. As shown in~\cref{tab:high_res_bench}, FALCON demonstrates superior performance across all these benchmarks.

To evaluate the performance of FALCON across a broader range of domains, we further conduct experiments with diverse popular benchmarks. As shown in~\cref{tab:general}, POPE~\cite{Li2023hallucination}, ScienceQA~\cite{lu2022learn}, MMBench~\cite{liu2025mmbench} and SEED-Bench~\cite{li2023seed} are selected to assess models' capabilities in hallucination, scientific reasoning, and comprehensive ability. FALCON achieves SOTA performance across all these benchmarks. The results demonstrate that FALCON exhibits superior capabilities across a wide range of tasks. 

\begin{table}[!t]
\caption{Evaluations on diverse high-resolution Benchmarks. ``$\textbf{V}^{*}_{Avg}$" represents the weighted average score for the two sub-tasks in the $V^{*}$ benchmark. \textdagger~denotes OCR inputs are utilized.}
\footnotesize
\label{tab:high_res_bench}
\centering
\setlength{\tabcolsep}{2pt}
\begin{tabular}{l c c c c}
    \toprule
    \textbf{Model} & \makecell{$\textbf{V}^{*}_{Avg}$} & \makecell{\textbf{DocVQA}} & \makecell{\textbf{TextVQA}} & \makecell{\textbf{ChartQA}}\\

    \midrule 
     \multicolumn{5}{c}{\textbf{Low-resolution MLLMs}} \\

    Qwen-VL~\cite{bai2023qwen}  &  -  & 65.1 & 63.8 & 59.0\\

    InstructBLIP~\cite{dai2023instructblip} & 34.0 & 4.5 & 50.7 & 10.9\\

    LLaVA-v1.5-7B~\cite{liu2024improved} & 48.7 & 8.5 & 58.2\textdagger & 17.8\\

    mPLUG-Owl2~\cite{ye2024mplug-owl2}  &  -  & 17.9 & 58.2\textdagger & 22.8 \\

    \midrule 
    \multicolumn{5}{c}{\textbf{High-resolution MLLMs}} \\

    UReader~\cite{ye2023ureader} &-& 65.4 & 57.6 & 59.3\\

    Monkey~\cite{li2024monkey} & - & 66.5  & 65.8 & 59.0\\

    LlaVA-FlexAttn~\cite{li2024flexattention} & 54.4  & - & 48.9 & -\\

    S2-Wrapper-7B~\cite{shi2024when_do}  & 55.5  & - & 61.0 & -\\

    LlaVA-Next-Vicuna7B~\cite{liu2024llavanext} & 61.8 & - & 64.9 & 54.3\\

    SliME-7B~\cite{zhang2024beyond} & - & - & 64.4  & 33.9\\

    \midrule

    \rowcolor[HTML]{faf0e6}
    FALCON & \textbf{64.4} & \textbf{74.4} & \textbf{66.5} & \textbf{68.8}\\
    \bottomrule
\end{tabular}
\vspace*{-3mm}
\end{table}

\begin{table}[!t]
\scriptsize
\centering
\caption{Ablation results on training data. We train FALCON with two-stage pipeline and exactly the same dataset as LLaVA-v1.5~\cite{liu2024improved}. The results confirm the superiority of the proposed method when the influence of training data is excluded. }
\label{tab:ablation_data}
\setlength{\tabcolsep}{1.5pt} 
\begin{tabular}{lccccccc}
    \toprule
    \textbf{Model} & \textbf{SQA} & \textbf{POPE} & \textbf{MMB} & \textbf{GQA} & \textbf{TextVQA} & \textbf{MM-Vet} & \textbf{SEED-Img}\\
    \midrule 
    
    LlaVA-v1.5-7B~\cite{liu2024improved} &  66.8 & 85.9 & 64.3 & 62.0 & 58.2 & 31.1 & 65.8\\
    
    \rowcolor[HTML]{faf0e6}
    Ours & \textbf{68.9} & \textbf{87.5} & \textbf{66.0} & \textbf{62.5} & \textbf{61.3} & \textbf{31.4} & \textbf{67.8}\\
    
    \bottomrule
\end{tabular}
\vspace*{-3mm}
\end{table}

\subsection{Ablation Study}
We perform extensive ablation studies to validate the effectiveness of the proposed ReCompact and ReAtten.

\begin{figure}[!t] 
    \centering
    \includegraphics[width=1\linewidth]{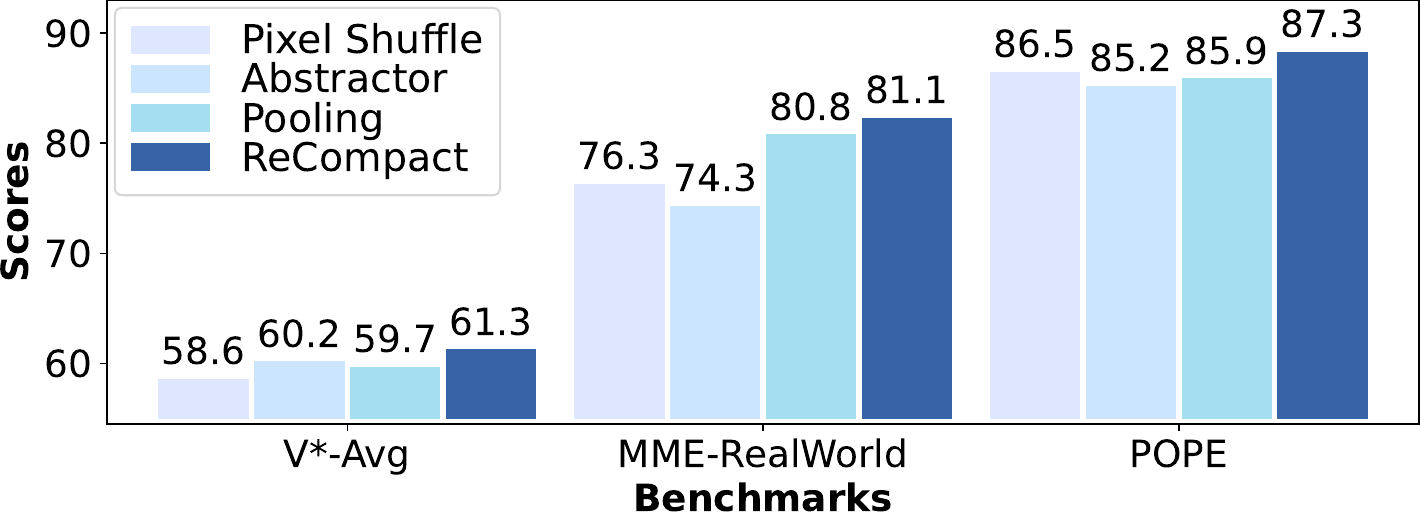}
    \caption{Comparison of different compression methods. We report the sum of the ``Avg-C" scores for the perception and reasoning tasks in MME-RealWorld. The superior performance demonstrates that ReCompact effectively eliminates redundancy and produces a more expressive and compact visual representation.}
    \label{fig:ablation_recompact}
    \vspace*{-6mm}
\end{figure}

\textbf{Ablation on Training Data}. To eliminate the influence of training data and demonstrate the advantages of the proposed method, we utilize exactly the same dataset as LLaVA-v1.5~\cite{liu2024improved} and employ a two-stage pipeline for training. As shown in~\cref{tab:ablation_data}, the proposed method exhibits significant performance advantages across multiple benchmarks when trained on the same dataset.

\textbf{Effectiveness of ReCompact mechanism}. To verify the advantages of the proposed ReCompact mechanism in reducing visual redundancy and compressing visual representations during the visual encoding, we compare the ReCompact with other commonly used compression methods. This includes pooling~\cite{yao2024deco,shi2024when_do}, Pixel Shuffle~\cite{chen2023internvl} and Abstractor~\cite{ye2024mplug-owl2}, where the Abstractor serves as a representative module that combines query tokens with cross-attention. 

As shown in~\cref{fig:ablation_recompact}, ReCompact demonstrates superior performance compared to other compression methods. These results indicate that ReCompact effectively eliminates visual redundancy and produces a more expressive compact visual representation. In contrast, we argue that the pooling layer and Pixel Shuffle merely reduces the number of tokens by merging adjacent ones, failing to eliminate redundancy that interferes with the LLM's understanding of visual content. Additionally, although Abstractor uses more parameters, its performance is generally lower than that of pooling and ReCompact. This result suggests that even millions of samples may be insufficient for fully training Abstractor, which diminishes their practical efficacy.

\begin{table}[!t]
\footnotesize
\centering
\caption{Evaluation on benchmarks across diverse domains. The POPE, SQA, MMBench and SEED-Bench are used to assess hallucination, scientific reasoning, and overall abilities, respectively.}
\label{tab:general}
\setlength{\tabcolsep}{4pt} 
\begin{tabular}{lcccc}
    \toprule
    \textbf{Model} & \textbf{SQA} & \textbf{POPE} & \textbf{MMB}&\textbf{SEED-Img}\\
    \midrule

    \multicolumn{5}{c}{\textbf{Low-Resolution MLLMs}} \\
    
    Qwen-VL~\cite{bai2023qwen} & 67.1 & - & 41.6 & 62.3\\
    InstructBLIP~\cite{dai2023instructblip} & 63.1 & 78.9 & 38.4 & 58.8\\
    LlaVA-v1.5-7B~\cite{liu2024improved} & 66.8 & 85.9 & 62.8& 65.8 \\
    mPLUG-Owl2~\cite{ye2024mplug-owl2} & 68.7 & 85.8 & 63.5& 64.5 \\
    \midrule 
    
    \multicolumn{5}{c}{\textbf{High-Resolution MLLMs}} \\
    
    Monkey~\cite{li2024monkey} & 69.4 & 83.7 & 57.4 & 64.3\\
    
    LlaVA-FlexAttn~\cite{li2024flexattention} & - & 85.9& 65.7&-\\
    
    S2-Wrapper-7B~\cite{shi2024when_do} & - & - & 66.2&67.9 \\
    
    LLaVA-Next-Vicuna7B~\cite{liu2024llavanext} & 70.1 & 86.5 & 66.5 & 69.6\\
    
    SliME-7B~\cite{zhang2024beyond} & 76.8 & 85.4 & 63.1 & 67.0\\

    \midrule 

    \rowcolor[HTML]{faf0e6}
    FALCON & \textbf{88.8} & \textbf{88.0} & \textbf{76.1}& \textbf{73.3}\\
    \bottomrule
\end{tabular}
\vspace*{-3mm}
\end{table}


\begin{table}[!t]
\scriptsize
\caption{Comparison of different methods for maintaining visual continuity. The ``Baseline" model refers to a condition where no specific methods are applied to preserve visual continuity. We report the ``Avg-C'' score for MME-RealWorld. The results show that the ReAtten module more effectively addresses the issue of visual fragmentation compared to other methods, thereby improving overall performance and reducing hallucinations.}	
\centering
\setlength{\tabcolsep}{6pt}
\begin{tabular}{c | c c c c }
    \toprule
    
    \multirow{2}{*}{\textbf{Method}} & \multirow{2}{*}{\makecell{$\textbf{V}^{*}_{Avg}$}} & \multicolumn{2}{c}{$\textbf{MME-RealWorld}$} & \multirow{2}{*}{\textbf{POPE}} \\
    
    \cmidrule(lr){3-4}
    
    & & \textbf{Perception} & \textbf{Reasoning} & \\

    \midrule

    Baseline & 51.3 & 38.2 & 35.6 & 85.7\\
    CIP~\cite{huang2024minimonkey} & 50.3 & 38.4 & 37.2 & 86.3 \\
    W-Atten~\cite{liu2024textmonkey} & 60.2 & 41.0 &  38.1 & 86.4\\
    
    \midrule 
    \rowcolor[HTML]{faf0e6}
    ReAtten & \textbf{61.3} & \textbf{42.1} & \textbf{39.0} & \textbf{87.3} \\

    \bottomrule
\end{tabular}
\label{tab:ablation_fragmentation}
\vspace*{-6mm}
\end{table}

\begin{figure*}[!t] 
    \centering
    \includegraphics[width=1\linewidth]{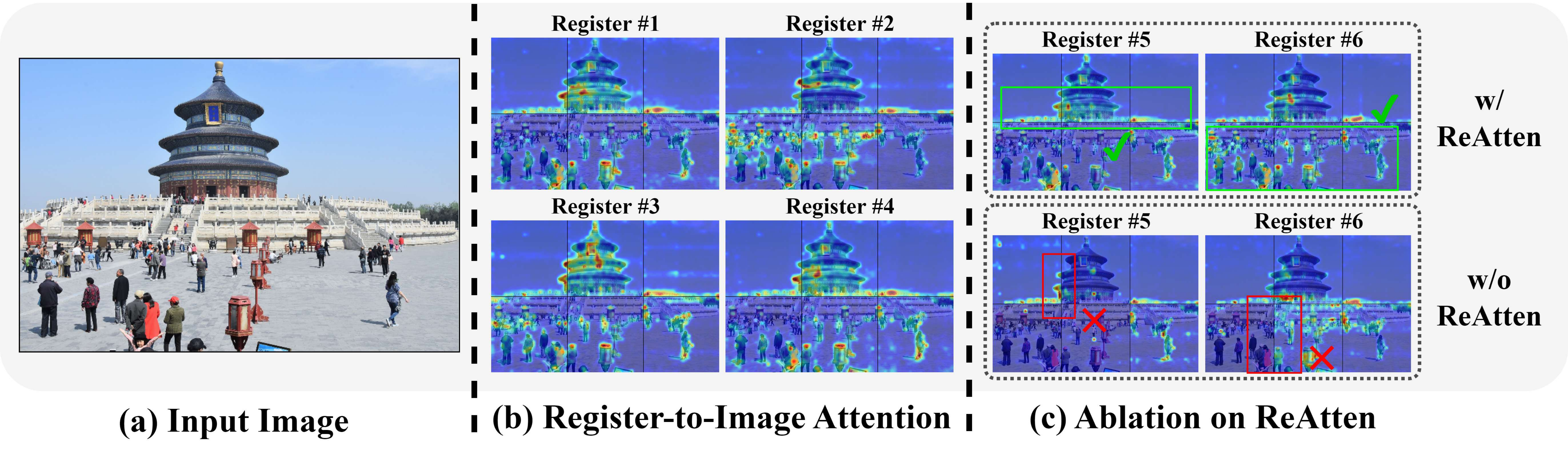}
    \caption{
        Visualization of the register-to-image attention map. We randomly select several visual registers and extract the attention scores that each register assigns to image tokens in the ViT model. We then create a heatmap and overlay it on the input image to visualize the attention patterns. The results in (b) shows that each visual register focuses on specific parts of the image, capturing rich visual patterns. Meanwhile, registers pay minimal attention to background areas, effectively avoiding the inclusion of redundancy. We also conduct qualitative ablation studies of the ReAtten module using visualization. As shown in (c), in the ViT model without ReAtten, the attention patterns across different sub-images appear extremely fragmented. In contrast, the ViT model with ReAtten shows continuous attention patterns, indicating effective information interaction between sub-images.
    }
    \label{fig:vis_reg_atten}
    \vspace*{-7mm}
\end{figure*}

\begin{figure}[!t] 
    \centering
    \includegraphics[width=0.9\linewidth]{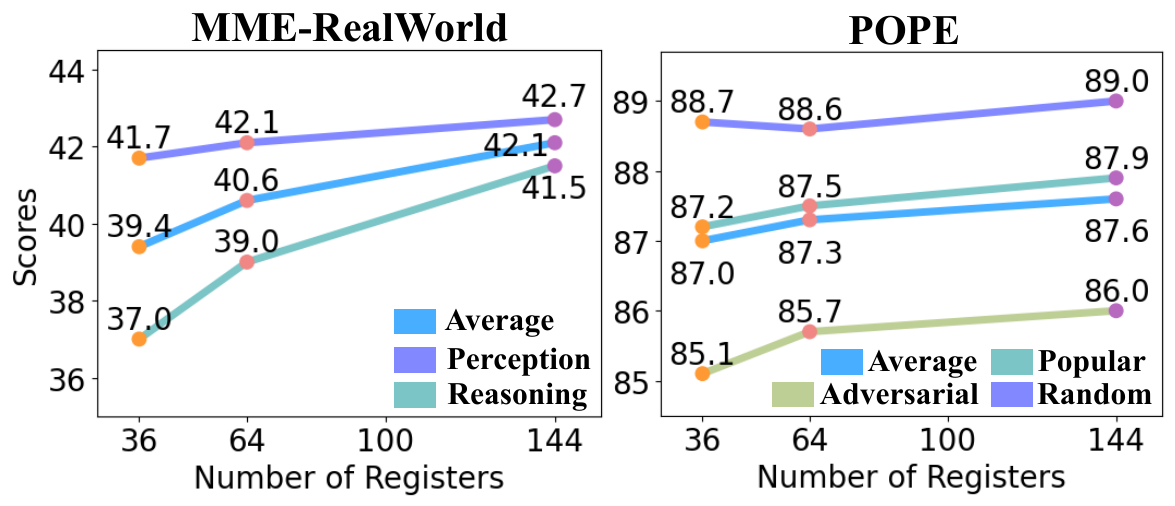}
    \caption{Ablation on different visual register counts. Increasing the number of visual registers generally enhances accuracy across both benchmarks. However, the improvement rate noticeably tapers off when the register count exceeds 64. Based on this observation, to better balance the efficiency and performance, we select register counts of 64 for FALCON.}
    \label{fig:ablation_reg}
    \vspace*{-7mm}
\end{figure}

\textbf{Effectiveness of ReAtten Module}. To validate the advantage of the proposed ReAtten Module, we compare ReAtten with other contemporary methods, including the Shifted Window Attention (W-Attn)~\cite{liu2021swin} adopted by TextMonkey~\cite{liu2024textmonkey} and the Complementary Image Pyramid (CIP) introduced by MiniMonkey~\cite{huang2024minimonkey}.

As shown in~\cref{tab:ablation_fragmentation}, the model incorporating ReAtten achieves significant improvements over the baseline across all benchmarks and demonstrates superiority compared to other methods. These results indicate that the ReAtten module effectively enhances the ability of the model in high-resolution perception and mitigates hallucinations by addressing the issue of visual fragmentation. In contrast, we argue that the information exchange in W-Atten is limited to a local window, which hinders sufficient information exchange among all sub-images. Meanwhile, CIP introduces further redundancy by repeatedly encoding the same visual content, which can even lead to a decline in performance.

\textbf{Ablation on the Number of Visual Registers}. To investigate the impact of changing the number of visual registers on model performance, we conduct experiments with visual register counts set to 36, 64, and 144. These counts correspond to visual token compression rates of $16\times$, $9\times$, and $4\times$, respectively. As shown in~\cref{fig:ablation_reg}, increasing the number of visual registers generally enhances the model's high-resolution comprehension capabilities and helps reduce hallucinations. However, when the register count exceeds 64, the improvement gains start tapering off noticeably. Meanwhile, increasing visual registers also demands more computational resources. Compared to the baseline model with 36 registers, models configured with 64 and 144 registers require 1.15 and 2.41 times more training time, respectively. Base on these observations, to achieve a better balance between efficiency and performance, we select 64 as the register counts for FALCON.

\subsection{Qualitative Analysis}
\label{sec:qualitative_analysis}

\textbf{Visualization and Analysis of Attention Maps}. 
To qualitatively analyze the proposed visual register technique, we visualize the register-to-image attention maps within ViT. Specifically, for a particular visual register, we extract the attention scores that this register assigns to image tokens in each sub-image. We then concatenate these attention scores across all sub-images, create a heatmap, and overlay it onto the input image. As shown in~\cref{fig:vis_reg_atten}(b), each visual register focuses on relatively distinct areas of the image, highlighting that it captures a diverse range of visual details. Moreover, the visual registers tend to avoid background areas, indicating their effectiveness in identifying and discarding visual redundancy. These properties enable visual registers to effectively produce a comprehensive and non-redundant visual representation. Additionally, to qualitatively evaluate the effectiveness of the ReAtten module, we visualize the attention maps of several particular registers within ViT models with and without ReAtten. As shown in~\cref{fig:vis_reg_atten}(c), in the ViT model without ReAtten, the attention patterns across different sub-images appear highly fragmented. In contrast, the ViT model with ReAtten exhibits continuous attention patterns, indicating effective information exchange between sub-images. 

\begin{figure*}[t] 
    \centering
    \includegraphics[width=0.925\linewidth]{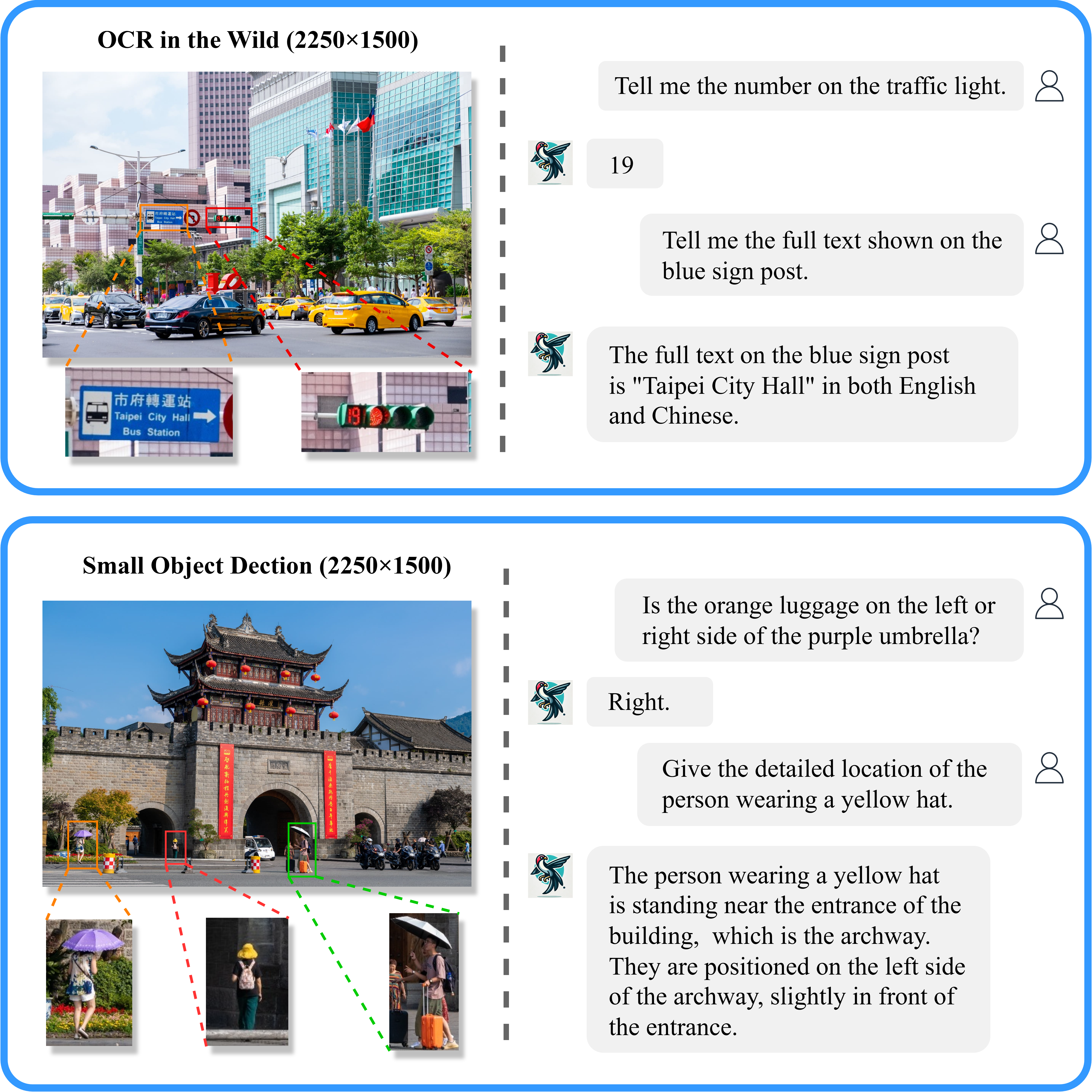}
    \caption{Qualitative analysis of FALCON in high-resolution images of natural scenes. The examples illustrate FALCON's exceptional capabilities in detecting small objects and recognizing small text in natural scenes, demonstrating its ability to capture rich fine-grained information in high-resolution images.}
    \label{fig:vqa_1}
    \vspace*{-5mm}
\end{figure*}

\begin{figure*}[t] 
    \centering
    \includegraphics[width=0.920\linewidth]{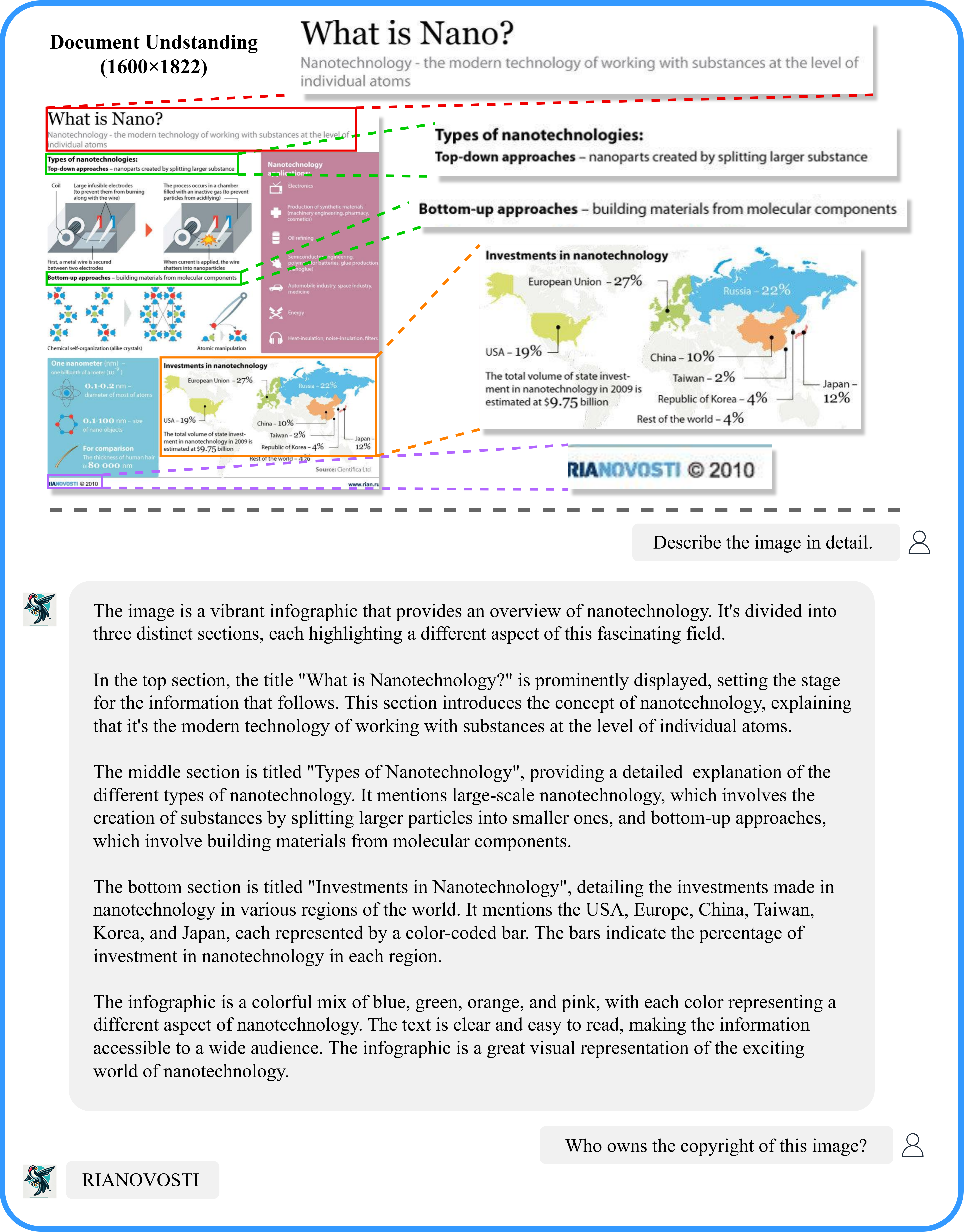}
    \caption{Qualitative analysis of FALCON in high-resolution images of documents. The example demonstrates FALCON's strong understanding and summarization capabilities for text-intensive, high-resolution document images. It also shows FALCON's ability to recognize small text. This proves FALCON's robustness in capturing rich, fine-grained information across different types of high-resolution images.}
    \label{fig:vqa_2}
\end{figure*}

\textbf{Case Study on Diverse Tasks}. As shown in~\cref{fig:vqa_1} and~\cref{fig:vqa_2}, we depict various examples to qualitatively validate the advanced high-resolution understanding abilities of FALCON.~\cref{fig:vqa_1} illustrates FALCON's exceptional ability to recognize small objects and text in natural scenes, demonstrating its capability to capture rich, fine-grained details in high-resolution images. Furthermore,~\cref{fig:vqa_2} highlights FALCON's proficiency in understanding and summarizing high-resolution document images with dense text, while also demonstrating its sensitivity to small text elements. These examples demonstrate FALCON's remarkable capabilities across various high-resolution vision-language tasks.

\section{Conclusion}
\label{sec:conclusion}

To address the visual redundancy and fragmentation in high-resolution MLLMs, we propose FALCON. FALCON employs an innovative visual register technique that simultaneously addresses both challenges. This technique uses a ReCompact mechanism to adaptively aggregate essential visual information through visual registers, creating a compact, non-redundant representation. Additionally, a novel ReAtten module is introduced to facilitate information exchange among sub-images via visual registers, thereby enhancing visual continuity during encoding. Extensive experiments demonstrate FALCON’s superiority in high-resolution understanding and validate the effectiveness of the proposed ReCompact and ReAtten.

{
    \small
    \bibliographystyle{ieeenat_fullname}
    \bibliography{main}

\begin{thebibliography}{66}
\providecommand{\natexlab}[1]{#1}
\providecommand{\url}[1]{\texttt{#1}}
\expandafter\ifx\csname urlstyle\endcsname\relax
  \providecommand{\doi}[1]{doi: #1}\else
  \providecommand{\doi}{doi: \begingroup \urlstyle{rm}\Url}\fi

\bibitem[Bai et~al.(2023)Bai, Bai, Yang, Wang, Tan, Wang, Lin, Zhou, and Zhou]{bai2023qwen}
Jinze Bai, Shuai Bai, Shusheng Yang, Shijie Wang, Sinan Tan, Peng Wang, Junyang Lin, Chang Zhou, and Jingren Zhou.
\newblock Qwen-vl: A versatile vision-language model for understanding, localization, text reading, and beyond.
\newblock \emph{arXiv preprint arXiv:2308.12966}, 1\penalty0 (2):\penalty0 3, 2023.

\bibitem[Chen et~al.(2024{\natexlab{a}})Chen, Shen, Shao, Deng, and Nie]{chen2024lion}
Gongwei Chen, Leyang Shen, Rui Shao, Xiang Deng, and Liqiang Nie.
\newblock Lion: Empowering multimodal large language model with dual-level visual knowledge.
\newblock In \emph{Proceedings of the IEEE/CVF Conference on Computer Vision and Pattern Recognition}, pages 26540--26550, 2024{\natexlab{a}}.

\bibitem[Chen et~al.(2025{\natexlab{a}})Chen, Zhou, Shao, Lyu, Zhou, Li, Li, Qi, and Nie]{chen2025SimpAgent}
Gongwei Chen, Xurui Zhou, Rui Shao, Yibo Lyu, Kaiwen Zhou, Wentao Li, Yinchuan Li, Zhongang Qi, and Liqiang Nie.
\newblock Less is more: Empowering gui agent with context-aware simplification.
\newblock In \emph{Proceedings of the IEEE/CVF International Conference on Computer Vision (ICCV)}, 2025{\natexlab{a}}.

\bibitem[Chen et~al.(2023)Chen, Zhu, Shen, Li, Liu, Zhang, Krishnamoorthi, Chandra, Xiong, and Elhoseiny]{chen2023minigpt}
Jun Chen, Deyao Zhu, Xiaoqian Shen, Xiang Li, Zechun Liu, Pengchuan Zhang, Raghuraman Krishnamoorthi, Vikas Chandra, Yunyang Xiong, and Mohamed Elhoseiny.
\newblock Minigpt-v2: large language model as a unified interface for vision-language multi-task learning.
\newblock \emph{arXiv preprint arXiv:2310.09478}, 2023.

\bibitem[Chen et~al.(2025{\natexlab{b}})Chen, Yuen, Xie, Yang, Chen, Wu, Yixing, Zhou, Liu, Wang, Zhou, Shao, Nie, Wang, HAO, Wang, and Shao]{chen2025spabench}
Jingxuan Chen, Derek Yuen, Bin Xie, Yuhao Yang, Gongwei Chen, Zhihao Wu, Li Yixing, Xurui Zhou, Weiwen Liu, Shuai Wang, Kaiwen Zhou, Rui Shao, Liqiang Nie, Yasheng Wang, Jianye HAO, Jun Wang, and Kun Shao.
\newblock Spa-bench: A comprehensive benchmark for smartphone agent evaluation.
\newblock In \emph{The Thirteenth International Conference on Learning Representations}, 2025{\natexlab{b}}.

\bibitem[Chen et~al.(2024{\natexlab{b}})Chen, Li, Dong, Zhang, He, Wang, Zhao, and Lin]{Lin2023ShareGPT4V}
Lin Chen, Jinsong Li, Xiaoyi Dong, Pan Zhang, Conghui He, Jiaqi Wang, Feng Zhao, and Dahua Lin.
\newblock Sharegpt4v: Improving large multi-modal models with better captions.
\newblock In \emph{European Conference on Computer Vision}, pages 370--387. Springer, 2024{\natexlab{b}}.

\bibitem[Chen et~al.(2024{\natexlab{c}})Chen, Wu, Wang, Su, Chen, Xing, Zhong, Zhang, Zhu, Lu, et~al.]{chen2023internvl}
Zhe Chen, Jiannan Wu, Wenhai Wang, Weijie Su, Guo Chen, Sen Xing, Muyan Zhong, Qinglong Zhang, Xizhou Zhu, Lewei Lu, et~al.
\newblock Internvl: Scaling up vision foundation models and aligning for generic visual-linguistic tasks.
\newblock In \emph{Proceedings of the IEEE/CVF conference on computer vision and pattern recognition}, pages 24185--24198, 2024{\natexlab{c}}.

\bibitem[Chiang et~al.(2023)Chiang, Li, Lin, Sheng, Wu, Zhang, Zheng, Zhuang, Zhuang, Gonzalez, et~al.]{chiang2023vicuna}
Wei-Lin Chiang, Zhuohan Li, Zi Lin, Ying Sheng, Zhanghao Wu, Hao Zhang, Lianmin Zheng, Siyuan Zhuang, Yonghao Zhuang, Joseph~E Gonzalez, et~al.
\newblock Vicuna: An open-source chatbot impressing gpt-4 with 90\%* chatgpt quality.
\newblock \emph{See https://vicuna. lmsys. org (accessed 14 April 2023)}, 2\penalty0 (3):\penalty0 6, 2023.

\bibitem[Dai et~al.(2023)Dai, Li, Li, Tiong, Zhao, Wang, Li, Fung, and Hoi]{dai2023instructblip}
Wenliang Dai, Junnan Li, Dongxu Li, Anthony Tiong, Junqi Zhao, Weisheng Wang, Boyang Li, Pascale Fung, and Steven Hoi.
\newblock Instruct{BLIP}: Towards general-purpose vision-language models with instruction tuning.
\newblock In \emph{Thirty-seventh Conference on Neural Information Processing Systems}, 2023.

\bibitem[Darcet et~al.(2024)Darcet, Oquab, Mairal, and Bojanowski]{darcet2024vision}
Timoth{\'e}e Darcet, Maxime Oquab, Julien Mairal, and Piotr Bojanowski.
\newblock Vision transformers need registers.
\newblock In \emph{The Twelfth International Conference on Learning Representations}, 2024.

\bibitem[Dosovitskiy et~al.(2021)Dosovitskiy, Beyer, Kolesnikov, Weissenborn, Zhai, Unterthiner, Dehghani, Minderer, Heigold, Gelly, Uszkoreit, and Houlsby]{dosovitskiy2021vit}
Alexey Dosovitskiy, Lucas Beyer, Alexander Kolesnikov, Dirk Weissenborn, Xiaohua Zhai, Thomas Unterthiner, Mostafa Dehghani, Matthias Minderer, Georg Heigold, Sylvain Gelly, Jakob Uszkoreit, and Neil Houlsby.
\newblock An image is worth 16x16 words: Transformers for image recognition at scale.
\newblock In \emph{International Conference on Learning Representations}, 2021.

\bibitem[Dubey et~al.(2024)Dubey, Jauhri, Pandey, Kadian, Al-Dahle, Letman, Mathur, Schelten, Yang, Fan, et~al.]{dubey2024llama3}
Abhimanyu Dubey, Abhinav Jauhri, Abhinav Pandey, Abhishek Kadian, Ahmad Al-Dahle, Aiesha Letman, Akhil Mathur, Alan Schelten, Amy Yang, Angela Fan, et~al.
\newblock The llama 3 herd of models.
\newblock \emph{arXiv preprint arXiv:2407.21783}, 2024.

\bibitem[Hendrycks and Gimpel(2016)]{hendrycks2016bridging}
Dan Hendrycks and Kevin Gimpel.
\newblock Bridging nonlinearities and stochastic regularizers with gaussian error linear units.
\newblock \emph{arXiv preprint arXiv:1606.08415}, 2016.

\bibitem[Hong et~al.(2024)Hong, Wang, Ding, Yu, Lv, Wang, Cheng, Huang, Ji, Xue, et~al.]{hong2024cogvlm2}
Wenyi Hong, Weihan Wang, Ming Ding, Wenmeng Yu, Qingsong Lv, Yan Wang, Yean Cheng, Shiyu Huang, Junhui Ji, Zhao Xue, et~al.
\newblock Cogvlm2: Visual language models for image and video understanding.
\newblock \emph{arXiv preprint arXiv:2408.16500}, 2024.

\bibitem[Hu et~al.(2024)Hu, Xu, Ye, Yan, Zhang, Zhang, Zhang, Jin, Huang, and Zhou]{hu2024mplug_docowl_1_5}
Anwen Hu, Haiyang Xu, Jiabo Ye, Ming Yan, Liang Zhang, Bo Zhang, Ji Zhang, Qin Jin, Fei Huang, and Jingren Zhou.
\newblock mplug-docowl 1.5: Unified structure learning for ocr-free document understanding.
\newblock In \emph{Findings of the Association for Computational Linguistics: EMNLP 2024}, pages 3096--3120, 2024.

\bibitem[Huang et~al.(2025{\natexlab{a}})Huang, Liu, Liang, Jin, and Bai]{huang2024minimonkey}
Mingxin Huang, Yuliang Liu, Dingkang Liang, Lianwen Jin, and Xiang Bai.
\newblock Mini-monkey: Alleviating the semantic sawtooth effect for lightweight {MLLM}s via complementary image pyramid.
\newblock In \emph{The Thirteenth International Conference on Learning Representations}, 2025{\natexlab{a}}.

\bibitem[Huang et~al.(2025{\natexlab{b}})Huang, Ding, Wang, Han, Liu, Zhao, Xu, Hou, Zhang, and Liang]{huang2024hires}
Runhui Huang, Xinpeng Ding, Chunwei Wang, Jianhua Han, Yulong Liu, Hengshuang Zhao, Hang Xu, Lu Hou, Wei Zhang, and Xiaodan Liang.
\newblock Hires-llava: Restoring fragmentation input in high-resolution large vision-language models.
\newblock In \emph{Proceedings of the Computer Vision and Pattern Recognition Conference}, pages 29814--29824, 2025{\natexlab{b}}.

\bibitem[Kim et~al.(2024)Kim, Pertsch, Karamcheti, Xiao, Balakrishna, Nair, Rafailov, Foster, Sanketi, Vuong, et~al.]{kimopenvla}
Moo~Jin Kim, Karl Pertsch, Siddharth Karamcheti, Ted Xiao, Ashwin Balakrishna, Suraj Nair, Rafael Rafailov, Ethan~P Foster, Pannag~R Sanketi, Quan Vuong, et~al.
\newblock Openvla: An open-source vision-language-action model.
\newblock In \emph{8th Annual Conference on Robot Learning}, 2024.

\bibitem[Li et~al.(2023{\natexlab{a}})Li, Wang, Wang, Ge, Ge, and Shan]{li2023seed}
Bohao Li, Rui Wang, Guangzhi Wang, Yuying Ge, Yixiao Ge, and Ying Shan.
\newblock Seed-bench: Benchmarking multimodal llms with generative comprehension.
\newblock \emph{arXiv preprint arXiv:2307.16125}, 2023{\natexlab{a}}.

\bibitem[Li et~al.(2025{\natexlab{a}})Li, Zhang, Guo, Zhang, Li, Zhang, Zhang, Zhang, Li, Liu, and Li]{li2024llava_onevision}
Bo Li, Yuanhan Zhang, Dong Guo, Renrui Zhang, Feng Li, Hao Zhang, Kaichen Zhang, Peiyuan Zhang, Yanwei Li, Ziwei Liu, and Chunyuan Li.
\newblock {LL}a{VA}-onevision: Easy visual task transfer.
\newblock \emph{Transactions on Machine Learning Research}, 2025{\natexlab{a}}.

\bibitem[Li et~al.(2025{\natexlab{b}})Li, Lv, Shao, Deng, Li, HAO, and Nie]{li2025star}
Hao Li, Qi Lv, Rui Shao, Xiang Deng, Yinchuan Li, Jianye HAO, and Liqiang Nie.
\newblock {STAR}: Learning diverse robot skill abstractions through rotation-augmented vector quantization.
\newblock In \emph{Forty-second International Conference on Machine Learning}, 2025{\natexlab{b}}.

\bibitem[Li et~al.(2023{\natexlab{b}})Li, Li, Savarese, and Hoi]{li2023blip2}
Junnan Li, Dongxu Li, Silvio Savarese, and Steven Hoi.
\newblock Blip-2: Bootstrapping language-image pre-training with frozen image encoders and large language models.
\newblock In \emph{International conference on machine learning}, pages 19730--19742. PMLR, 2023{\natexlab{b}}.

\bibitem[Li et~al.(2024{\natexlab{a}})Li, Chen, Cai, Chen, Hong, Chen, Shen, and Gan]{li2024flexattention}
Junyan Li, Delin Chen, Tianle Cai, Peihao Chen, Yining Hong, Zhenfang Chen, Yikang Shen, and Chuang Gan.
\newblock Flexattention for efficient high-resolution vision-language models.
\newblock In \emph{European Conference on Computer Vision}, pages 286--302, 2024{\natexlab{a}}.

\bibitem[Li et~al.(2025{\natexlab{c}})Li, Hu, Shao, Shen, and Nie]{li2025lion}
Wei Li, Bing Hu, Rui Shao, Leyang Shen, and Liqiang Nie.
\newblock Lion-fs: Fast \& slow video-language thinker as online video assistant.
\newblock In \emph{Proceedings of the Computer Vision and Pattern Recognition Conference}, pages 3240--3251, 2025{\natexlab{c}}.

\bibitem[Li et~al.(2023{\natexlab{c}})Li, Du, Zhou, Wang, Zhao, and Wen]{Li2023hallucination}
Yifan Li, Yifan Du, Kun Zhou, Jinpeng Wang, Xin Zhao, and Ji-Rong Wen.
\newblock Evaluating object hallucination in large vision-language models.
\newblock In \emph{The 2023 Conference on Empirical Methods in Natural Language Processing}, 2023{\natexlab{c}}.

\bibitem[Li et~al.(2024{\natexlab{b}})Li, Zhang, Wang, Zhong, Chen, Chu, Liu, and Jia]{li2024mini_gemini}
Yanwei Li, Yuechen Zhang, Chengyao Wang, Zhisheng Zhong, Yixin Chen, Ruihang Chu, Shaoteng Liu, and Jiaya Jia.
\newblock Mini-gemini: Mining the potential of multi-modality vision language models.
\newblock \emph{arXiv preprint arXiv:2403.18814}, 2024{\natexlab{b}}.

\bibitem[Li et~al.(2024{\natexlab{c}})Li, Xie, Shao, Chen, Jiang, and Nie]{li2024optimus}
Zaijing Li, Yuquan Xie, Rui Shao, Gongwei Chen, Dongmei Jiang, and Liqiang Nie.
\newblock Optimus-1: Hybrid multimodal memory empowered agents excel in long-horizon tasks.
\newblock In \emph{Advances in neural information processing systems}, 2024{\natexlab{c}}.

\bibitem[Li et~al.(2024{\natexlab{d}})Li, Yang, Liu, Ma, Zhang, Yang, Sun, Liu, and Bai]{li2024monkey}
Zhang Li, Biao Yang, Qiang Liu, Zhiyin Ma, Shuo Zhang, Jingxu Yang, Yabo Sun, Yuliang Liu, and Xiang Bai.
\newblock Monkey: Image resolution and text label are important things for large multi-modal models.
\newblock In \emph{Proceedings of the IEEE/CVF Conference on Computer Vision and Pattern Recognition}, pages 26763--26773, 2024{\natexlab{d}}.

\bibitem[Li et~al.(2025{\natexlab{d}})Li, Xie, Shao, Chen, Jiang, and Nie]{li2025optimus2}
Zaijing Li, Yuquan Xie, Rui Shao, Gongwei Chen, Dongmei Jiang, and Liqiang Nie.
\newblock Optimus-2: Multimodal minecraft agent with goal-observation-action conditioned policy.
\newblock In \emph{2025 IEEE/CVF Conference on Computer Vision and Pattern Recognition (CVPR)}. IEEE, 2025{\natexlab{d}}.

\bibitem[Liu et~al.(2024{\natexlab{a}})Liu, Li, Li, and Lee]{liu2024improved}
Haotian Liu, Chunyuan Li, Yuheng Li, and Yong~Jae Lee.
\newblock Improved baselines with visual instruction tuning.
\newblock In \emph{Proceedings of the IEEE/CVF Conference on Computer Vision and Pattern Recognition}, pages 26296--26306, 2024{\natexlab{a}}.

\bibitem[Liu et~al.(2024{\natexlab{b}})Liu, Li, Li, Li, Zhang, Shen, and Lee]{liu2024llavanext}
Haotian Liu, Chunyuan Li, Yuheng Li, Bo Li, Yuanhan Zhang, Sheng Shen, and Yong~Jae Lee.
\newblock Llava-next: Improved reasoning, ocr, and world knowledge, 2024{\natexlab{b}}.

\bibitem[Liu et~al.(2024{\natexlab{c}})Liu, Li, Wu, and Lee]{liu2024visual}
Haotian Liu, Chunyuan Li, Qingyang Wu, and Yong~Jae Lee.
\newblock Visual instruction tuning.
\newblock \emph{Advances in neural information processing systems}, 36, 2024{\natexlab{c}}.

\bibitem[Liu et~al.(2024{\natexlab{d}})Liu, Yang, Liu, Li, Ma, Zhang, and Bai]{liu2024textmonkey}
Yuliang Liu, Biao Yang, Qiang Liu, Zhang Li, Zhiyin Ma, Shuo Zhang, and Xiang Bai.
\newblock Textmonkey: An ocr-free large multimodal model for understanding document.
\newblock \emph{arXiv preprint arXiv:2403.04473}, 2024{\natexlab{d}}.

\bibitem[Liu et~al.(2025)Liu, Duan, Zhang, Li, Zhang, Zhao, Yuan, Wang, He, Liu, et~al.]{liu2025mmbench}
Yuan Liu, Haodong Duan, Yuanhan Zhang, Bo Li, Songyang Zhang, Wangbo Zhao, Yike Yuan, Jiaqi Wang, Conghui He, Ziwei Liu, et~al.
\newblock Mmbench: Is your multi-modal model an all-around player?
\newblock In \emph{European Conference on Computer Vision}, pages 216--233. Springer, 2025.

\bibitem[Liu et~al.(2021)Liu, Lin, Cao, Hu, Wei, Zhang, Lin, and Guo]{liu2021swin}
Ze Liu, Yutong Lin, Yue Cao, Han Hu, Yixuan Wei, Zheng Zhang, Stephen Lin, and Baining Guo.
\newblock Swin transformer: Hierarchical vision transformer using shifted windows.
\newblock In \emph{Proceedings of the IEEE/CVF international conference on computer vision}, pages 10012--10022, 2021.

\bibitem[Lu et~al.(2024)Lu, Liu, Zhang, Wang, Dong, Liu, Sun, Ren, Li, Sun, et~al.]{lu2024deepseek}
Haoyu Lu, Wen Liu, Bo Zhang, Bingxuan Wang, Kai Dong, Bo Liu, Jingxiang Sun, Tongzheng Ren, Zhuoshu Li, Yaofeng Sun, et~al.
\newblock Deepseek-vl: towards real-world vision-language understanding.
\newblock \emph{arXiv preprint arXiv:2403.05525}, 2024.

\bibitem[Lu et~al.(2022)Lu, Mishra, Xia, Qiu, Chang, Zhu, Tafjord, Clark, and Kalyan]{lu2022learn}
Pan Lu, Swaroop Mishra, Tony Xia, Liang Qiu, Kai-Wei Chang, Song-Chun Zhu, Oyvind Tafjord, Peter Clark, and Ashwin Kalyan.
\newblock Learn to explain: Multimodal reasoning via thought chains for science question answering.
\newblock In \emph{The 36th Conference on Neural Information Processing Systems (NeurIPS)}, 2022.

\bibitem[Luo et~al.(2024)Luo, Zhou, Zhang, Zheng, Sun, and Ji]{luo2024feast}
Gen Luo, Yiyi Zhou, Yuxin Zhang, Xiawu Zheng, Xiaoshuai Sun, and Rongrong Ji.
\newblock Feast your eyes: Mixture-of-resolution adaptation for multimodal large language models.
\newblock \emph{arXiv preprint arXiv:2403.03003}, 2024.

\bibitem[Lv et~al.(2025)Lv, Li, Deng, Shao, Li, Hao, Gao, Wang, and Nie]{lv2025spatial}
Qi Lv, Hao Li, Xiang Deng, Rui Shao, Yinchuan Li, Jianye Hao, Longxiang Gao, Michael~Yu Wang, and Liqiang Nie.
\newblock Spatial-temporal graph diffusion policy with kinematic modeling for bimanual robotic manipulation.
\newblock In \emph{Proceedings of the Computer Vision and Pattern Recognition Conference}, pages 17394--17404, 2025.

\bibitem[Masry et~al.(2022)Masry, Do, Tan, Joty, and Hoque]{masry2022chartqa}
Ahmed Masry, Xuan~Long Do, Jia~Qing Tan, Shafiq Joty, and Enamul Hoque.
\newblock Chartqa: A benchmark for question answering about charts with visual and logical reasoning.
\newblock In \emph{Findings of the Association for Computational Linguistics: ACL 2022}, pages 2263--2279, 2022.

\bibitem[Mathew et~al.(2021)Mathew, Karatzas, and Jawahar]{mathew2021docvqa}
Minesh Mathew, Dimosthenis Karatzas, and CV Jawahar.
\newblock Docvqa: A dataset for vqa on document images.
\newblock In \emph{Proceedings of the IEEE/CVF winter conference on applications of computer vision}, pages 2200--2209, 2021.

\bibitem[Radford et~al.(2021)Radford, Kim, Hallacy, Ramesh, Goh, Agarwal, Sastry, Askell, Mishkin, Clark, et~al.]{radford2021learning}
Alec Radford, Jong~Wook Kim, Chris Hallacy, Aditya Ramesh, Gabriel Goh, Sandhini Agarwal, Girish Sastry, Amanda Askell, Pamela Mishkin, Jack Clark, et~al.
\newblock Learning transferable visual models from natural language supervision.
\newblock In \emph{International conference on machine learning}, pages 8748--8763. PMLR, 2021.

\bibitem[Shao et~al.(2019)Shao, Lan, Li, and Yuen]{shao2019multi}
Rui Shao, Xiangyuan Lan, Jiawei Li, and Pong~C Yuen.
\newblock Multi-adversarial discriminative deep domain generalization for face presentation attack detection.
\newblock In \emph{Proceedings of the IEEE/CVF conference on computer vision and pattern recognition}, pages 10023--10031, 2019.

\bibitem[Shao et~al.(2023)Shao, Wu, and Liu]{shao2023detecting}
Rui Shao, Tianxing Wu, and Ziwei Liu.
\newblock Detecting and grounding multi-modal media manipulation.
\newblock In \emph{Proceedings of the IEEE/CVF Conference on Computer Vision and Pattern Recognition}, pages 6904--6913, 2023.

\bibitem[Shao et~al.(2024)Shao, Wu, Wu, Nie, and Liu]{shao2024detecting}
Rui Shao, Tianxing Wu, Jianlong Wu, Liqiang Nie, and Ziwei Liu.
\newblock Detecting and grounding multi-modal media manipulation and beyond.
\newblock \emph{IEEE Transactions on Pattern Analysis and Machine Intelligence}, 2024.

\bibitem[Shen et~al.(2024)Shen, Chen, Shao, Guan, and Nie]{shen2024mome}
Leyang Shen, Gongwei Chen, Rui Shao, Weili Guan, and Liqiang Nie.
\newblock Mome: Mixture of multimodal experts for generalist multimodal large language models.
\newblock In \emph{Advances in neural information processing systems}, 2024.

\bibitem[Shi et~al.(2024)Shi, Wu, Mao, Wang, and Darrell]{shi2024when_do}
Baifeng Shi, Ziyang Wu, Maolin Mao, Xin Wang, and Trevor Darrell.
\newblock When do we not need larger vision models?
\newblock In \emph{European Conference on Computer Vision}, pages 444--462, 2024.

\bibitem[Singh et~al.(2019)Singh, Natarajan, Shah, Jiang, Chen, Batra, Parikh, and Rohrbach]{singh2019textvqa}
Amanpreet Singh, Vivek Natarajan, Meet Shah, Yu Jiang, Xinlei Chen, Dhruv Batra, Devi Parikh, and Marcus Rohrbach.
\newblock Towards vqa models that can read.
\newblock In \emph{Proceedings of the IEEE/CVF conference on computer vision and pattern recognition}, pages 8317--8326, 2019.

\bibitem[Team et~al.(2023)Team, Anil, Borgeaud, Alayrac, Yu, Soricut, Schalkwyk, Dai, Hauth, Millican, et~al.]{team2023gemini}
Gemini Team, Rohan Anil, Sebastian Borgeaud, Jean-Baptiste Alayrac, Jiahui Yu, Radu Soricut, Johan Schalkwyk, Andrew~M Dai, Anja Hauth, Katie Millican, et~al.
\newblock Gemini: a family of highly capable multimodal models.
\newblock \emph{arXiv preprint arXiv:2312.11805}, 2023.

\bibitem[Touvron et~al.(2023{\natexlab{a}})Touvron, Lavril, Izacard, Martinet, Lachaux, Lacroix, Rozi{\`e}re, Goyal, Hambro, Azhar, et~al.]{touvron2023llama}
Hugo Touvron, Thibaut Lavril, Gautier Izacard, Xavier Martinet, Marie-Anne Lachaux, Timoth{\'e}e Lacroix, Baptiste Rozi{\`e}re, Naman Goyal, Eric Hambro, Faisal Azhar, et~al.
\newblock Llama: Open and efficient foundation language models.
\newblock \emph{arXiv preprint arXiv:2302.13971}, 2023{\natexlab{a}}.

\bibitem[Touvron et~al.(2023{\natexlab{b}})Touvron, Martin, Stone, Albert, Almahairi, Babaei, Bashlykov, Batra, Bhargava, Bhosale, et~al.]{touvron2023llama2}
Hugo Touvron, Louis Martin, Kevin Stone, Peter Albert, Amjad Almahairi, Yasmine Babaei, Nikolay Bashlykov, Soumya Batra, Prajjwal Bhargava, Shruti Bhosale, et~al.
\newblock Llama 2: Open foundation and fine-tuned chat models.
\newblock \emph{arXiv preprint arXiv:2307.09288}, 2023{\natexlab{b}}.

\bibitem[Wu and Xie(2024)]{wu2024vstar}
Penghao Wu and Saining Xie.
\newblock V*: Guided visual search as a core mechanism in multimodal llms.
\newblock In \emph{Proceedings of the IEEE/CVF Conference on Computer Vision and Pattern Recognition}, pages 13084--13094, 2024.

\bibitem[Xie et~al.(2025)Xie, Shao, Chen, Zhou, Li, Liu, Zhang, and Nie]{xie2025gui}
Bin Xie, Rui Shao, Gongwei Chen, Kaiwen Zhou, Yinchuan Li, Jie Liu, Min Zhang, and Liqiang Nie.
\newblock Gui-explorer: Autonomous exploration and mining of transition-aware knowledge for gui agent.
\newblock In \emph{Annual Meeting of the Association for Computational Linguistics (ACL)}, 2025.

\bibitem[Yao et~al.(2024{\natexlab{a}})Yao, Li, Ren, Wang, Liu, Sun, and Hou]{yao2024deco}
Linli Yao, Lei Li, Shuhuai Ren, Lean Wang, Yuanxin Liu, Xu Sun, and Lu Hou.
\newblock Deco: Decoupling token compression from semantic abstraction in multimodal large language models.
\newblock \emph{arXiv preprint arXiv:2405.20985}, 2024{\natexlab{a}}.

\bibitem[Yao et~al.(2024{\natexlab{b}})Yao, Yu, Zhang, Wang, Cui, Zhu, Cai, Li, Zhao, He, et~al.]{yao2024minicpm}
Yuan Yao, Tianyu Yu, Ao Zhang, Chongyi Wang, Junbo Cui, Hongji Zhu, Tianchi Cai, Haoyu Li, Weilin Zhao, Zhihui He, et~al.
\newblock Minicpm-v: A gpt-4v level mllm on your phone.
\newblock \emph{arXiv preprint arXiv:2408.01800}, 2024{\natexlab{b}}.

\bibitem[Ye et~al.(2023{\natexlab{a}})Ye, Hu, Xu, Ye, Yan, Dan, Zhao, Xu, Li, Tian, et~al.]{ye2023mplug_docowl}
Jiabo Ye, Anwen Hu, Haiyang Xu, Qinghao Ye, Ming Yan, Yuhao Dan, Chenlin Zhao, Guohai Xu, Chenliang Li, Junfeng Tian, et~al.
\newblock mplug-docowl: Modularized multimodal large language model for document understanding.
\newblock \emph{arXiv preprint arXiv:2307.02499}, 2023{\natexlab{a}}.

\bibitem[Ye et~al.(2023{\natexlab{b}})Ye, Hu, Xu, Ye, Yan, Xu, Li, Tian, Qian, Zhang, et~al.]{ye2023ureader}
Jiabo Ye, Anwen Hu, Haiyang Xu, Qinghao Ye, Ming Yan, Guohai Xu, Chenliang Li, Junfeng Tian, Qi Qian, Ji Zhang, et~al.
\newblock Ureader: Universal ocr-free visually-situated language understanding with multimodal large language model.
\newblock In \emph{Findings of the Association for Computational Linguistics: EMNLP 2023}, pages 2841--2858, 2023{\natexlab{b}}.

\bibitem[Ye et~al.(2023{\natexlab{c}})Ye, Xu, Xu, Ye, Yan, Zhou, Wang, Hu, Shi, Shi, et~al.]{ye2023mplug}
Qinghao Ye, Haiyang Xu, Guohai Xu, Jiabo Ye, Ming Yan, Yiyang Zhou, Junyang Wang, Anwen Hu, Pengcheng Shi, Yaya Shi, et~al.
\newblock mplug-owl: Modularization empowers large language models with multimodality.
\newblock \emph{arXiv preprint arXiv:2304.14178}, 2023{\natexlab{c}}.

\bibitem[Ye et~al.(2024)Ye, Xu, Ye, Yan, Hu, Liu, Qian, Zhang, and Huang]{ye2024mplug-owl2}
Qinghao Ye, Haiyang Xu, Jiabo Ye, Ming Yan, Anwen Hu, Haowei Liu, Qi Qian, Ji Zhang, and Fei Huang.
\newblock mplug-owl2: Revolutionizing multi-modal large language model with modality collaboration.
\newblock In \emph{Proceedings of the IEEE/CVF Conference on Computer Vision and Pattern Recognition}, pages 13040--13051, 2024.

\bibitem[Yu et~al.(2016)Yu, Poirson, Yang, Berg, and Berg]{yu2016modeling}
Licheng Yu, Patrick Poirson, Shan Yang, Alexander~C Berg, and Tamara~L Berg.
\newblock Modeling context in referring expressions.
\newblock In \emph{Computer Vision--ECCV 2016: 14th European Conference, Amsterdam, The Netherlands, October 11-14, 2016, Proceedings, Part II 14}, pages 69--85. Springer, 2016.

\bibitem[Zhai et~al.(2023)Zhai, Mustafa, Kolesnikov, and Beyer]{zhai2023sigmoid}
Xiaohua Zhai, Basil Mustafa, Alexander Kolesnikov, and Lucas Beyer.
\newblock Sigmoid loss for language image pre-training.
\newblock In \emph{Proceedings of the IEEE/CVF International Conference on Computer Vision}, pages 11975--11986, 2023.

\bibitem[Zhang et~al.(2024{\natexlab{a}})Zhang, Dong, Zang, Cao, Qian, Chen, Guo, Duan, Wang, Ouyang, et~al.]{zhang2024ixc2_5}
Pan Zhang, Xiaoyi Dong, Yuhang Zang, Yuhang Cao, Rui Qian, Lin Chen, Qipeng Guo, Haodong Duan, Bin Wang, Linke Ouyang, et~al.
\newblock Internlm-xcomposer-2.5: A versatile large vision language model supporting long-contextual input and output.
\newblock \emph{arXiv preprint arXiv:2407.03320}, 2024{\natexlab{a}}.

\bibitem[Zhang et~al.(2024{\natexlab{b}})Zhang, Lyu, Shao, Chen, Guan, and Nie]{zhang2024token}
Renshan Zhang, Yibo Lyu, Rui Shao, Gongwei Chen, Weili Guan, and Liqiang Nie.
\newblock Token-level correlation-guided compression for efficient multimodal document understanding.
\newblock \emph{arXiv preprint arXiv:2407.14439}, 2024{\natexlab{b}}.

\bibitem[Zhang et~al.(2025{\natexlab{a}})Zhang, Quan, Shen, Yuan, Yan, Xie, Wang, Gu, Tang, and Ye]{zhang2024redundancy}
Xiaofeng Zhang, Yihao Quan, Chen Shen, Xiaosong Yuan, Shaotian Yan, Liang Xie, Wenxiao Wang, Chaochen Gu, Hao Tang, and Jieping Ye.
\newblock From redundancy to relevance: Enhancing explainability in multimodal large language models.
\newblock \emph{Annual Conference of the Nations of the Americas Chapter of the Association for Computational Linguistics}, 2025{\natexlab{a}}.

\bibitem[Zhang et~al.(2025{\natexlab{b}})Zhang, Zhang, Tian, Fu, Zhang, Wu, Li, Wang, Wen, Zhang, Wang, and Jin]{zhang2024mme_realword}
YiFan Zhang, Huanyu Zhang, Haochen Tian, Chaoyou Fu, Shuangqing Zhang, Junfei Wu, Feng Li, Kun Wang, Qingsong Wen, Zhang Zhang, Liang Wang, and Rong Jin.
\newblock {MME}-realworld: Could your multimodal {LLM} challenge high-resolution real-world scenarios that are difficult for humans?
\newblock In \emph{The Thirteenth International Conference on Learning Representations}, 2025{\natexlab{b}}.

\bibitem[Zhang et~al.(2024{\natexlab{c}})Zhang, Wen, Fu, Wang, Zhang, Wang, and Jin]{zhang2024beyond}
Yi-Fan Zhang, Qingsong Wen, Chaoyou Fu, Xue Wang, Zhang Zhang, Liang Wang, and Rong Jin.
\newblock Beyond llava-hd: Diving into high-resolution large multimodal models.
\newblock \emph{arXiv preprint arXiv:2406.08487}, 2024{\natexlab{c}}.

\end{thebibliography}
}


\end{document}